\documentclass{article} 
\usepackage{microtype}
\usepackage{graphicx}
\usepackage{subfigure}
\usepackage{booktabs} 

\PassOptionsToPackage{hyphens}{url}
\usepackage{xr-hyper}
\usepackage{hyperref}
\usepackage[dvipsnames]{xcolor}
\definecolor{myblue}{rgb}{0,0.22,0.45}
\hypersetup{
    colorlinks,
    linkcolor={black},
    citecolor={myblue},
    urlcolor={myblue}
}

\usepackage{amsmath}
\usepackage{amssymb}
\usepackage{mathtools}
\usepackage{amsthm}
\usepackage{microtype}
\usepackage{subfigure}
\makeatletter
\renewcommand{\thesubfigure}{\alph{subfigure}}
\renewcommand{\@thesubfigure}{(\thesubfigure)\hskip\subfiglabelskip}
\makeatother
\usepackage{wrapfig}
\usepackage{booktabs} 
\usepackage{float} 
\usepackage{xspace}

\usepackage[capitalize,noabbrev]{cleveref}

\definecolor{myred}{rgb}{0.8,0,0}
\definecolor{laetigreen}{rgb}{0.06,0.6,0.15}

\newcommand{\ie}{i.e.\,}
\newcommand{\eg}{e.g.\,}

\usepackage{collas2023_conference,times}
\usepackage{easyReview}

\usepackage[most]{tcolorbox}

\definecolor{linequote}{RGB}{224,215,188}
\definecolor{backquote}{RGB}{249,245,233}
\newcommand{\fon}[1]{\fontfamily{#1}\selectfont}
\newtcolorbox{myquote}[1][]{%
    enhanced, breakable, 
    size=minimal,
    frame hidden, boxrule=0pt,
    sharp corners,
    colback=backquote,
    #1
}

\title{Augmenting Autotelic Agents with \\Large Language Models}

\author{Cédric Colas \\
MIT, Inria \\
\texttt{ccolas@mit.edu} \\
\And 
Laetitia Teodorescu  \\
Inria \\
\texttt{laetitia.teodorescu@inria.fr} \\
\AND
Pierre-Yves Oudeyer  \\
Inria \\
\And 
Xingdi Yuan  \\
Microsoft Research \\
\And 
Marc-Alexandre Côté  \\
Microsoft Research \\
}

\preprintcopy 

\begin{document}

\maketitle

\begin{abstract}
Humans learn to master open-ended repertoires of skills by imagining and practicing their own goals. This \textit{autotelic learning process}, literally the pursuit of self-generated (\textit{auto}) goals (\textit{telos}), becomes more and more open-ended as the goals become more diverse, abstract and creative. The resulting exploration of the space of possible skills is supported by an inter-individual exploration: goal representations are culturally evolved and transmitted across individuals, in particular using language. 
Current artificial agents mostly rely on predefined goal representations corresponding to goal spaces that are either bounded (\eg list of instructions), or unbounded (\eg the space of possible visual inputs) but are rarely endowed with the ability to reshape their goal representations, to form new abstractions or to imagine creative goals. 
In this paper, we introduce a \textit{language model augmented autotelic agent} (LMA3) that leverages a pretrained language model (LM) to support the representation, generation and learning of diverse, abstract, human-relevant goals. 
The LM is used as an imperfect model of human cultural transmission; an attempt to capture aspects of humans' common-sense, intuitive physics and overall interests. Specifically, it supports three key components of the autotelic architecture: 1)~a relabeler that describes the goals achieved in the agent's trajectories, 2)~a goal generator that suggests new high-level goals along with their decomposition into subgoals the agent already masters, and 3)~reward functions for each of these goals. Without relying on any hand-coded goal representations, reward functions or curriculum, we show that LMA3 agents learn to master a large diversity of skills in a task-agnostic text-based environment.
\end{abstract}

\section{Introduction} 

Each human learns an open-ended set of skills across their life: from throwing objects, building Lego structures and drawing stick figures, to perhaps playing tennis professionally, building bridges or conveying emotions through paintings. These skills are not the direct product of evolution but the result of goal-directed learning processes. Although most living creatures pursue goals that directly impact their survival, humans seem to spend most of their time pursuing frivolous goals: \eg watching movies, playing video games, or taking photographs \citep{chu2020play}.

The field of developmental AI models these evolved tendencies with \textit{intrinsic motivations} (IM), internal reward systems that drive agents to experience interesting situations and explore their environment \citep{singh2010intrinsically,oudeyer2009intrinsic}. While knowledge-based IMs drive agents to learn about the world \citep{aubret2019survey,linke2020adapting}, competence-based IMs drive agents to learn to control their environment  \citep{oudeyer2009intrinsic,colas2022autotelic}. Agents endowed with these intrinsic motivations are \textit{autotelic}; they are intrinsically driven (\textit{auto}) to learn to represent, generate, pursue and master their own goals (\textit{telos}) \citep{colas2022autotelic}. Open-ended learning processes require the joint training of a problem generator (\eg environment dynamics, opponents, goals) and a problem solver: the former challenging the latter in more and more complex scenarios, providing a never-ending curriculum for the problem solver \citep{schmidhuber2013powerplay, wang2020enhanced, ecoffet2021first, jiang2021prioritized, team2023human}. Autotelic agents are specifically designed for open-ended \textit{skill} learning by jointly training a \textit{goal} generator and a \textit{goal-conditioned} policy, see a review in \cite{colas2022autotelic}. 

\begin{figure}[t!]
    \centering
    \includegraphics[width=1\linewidth]{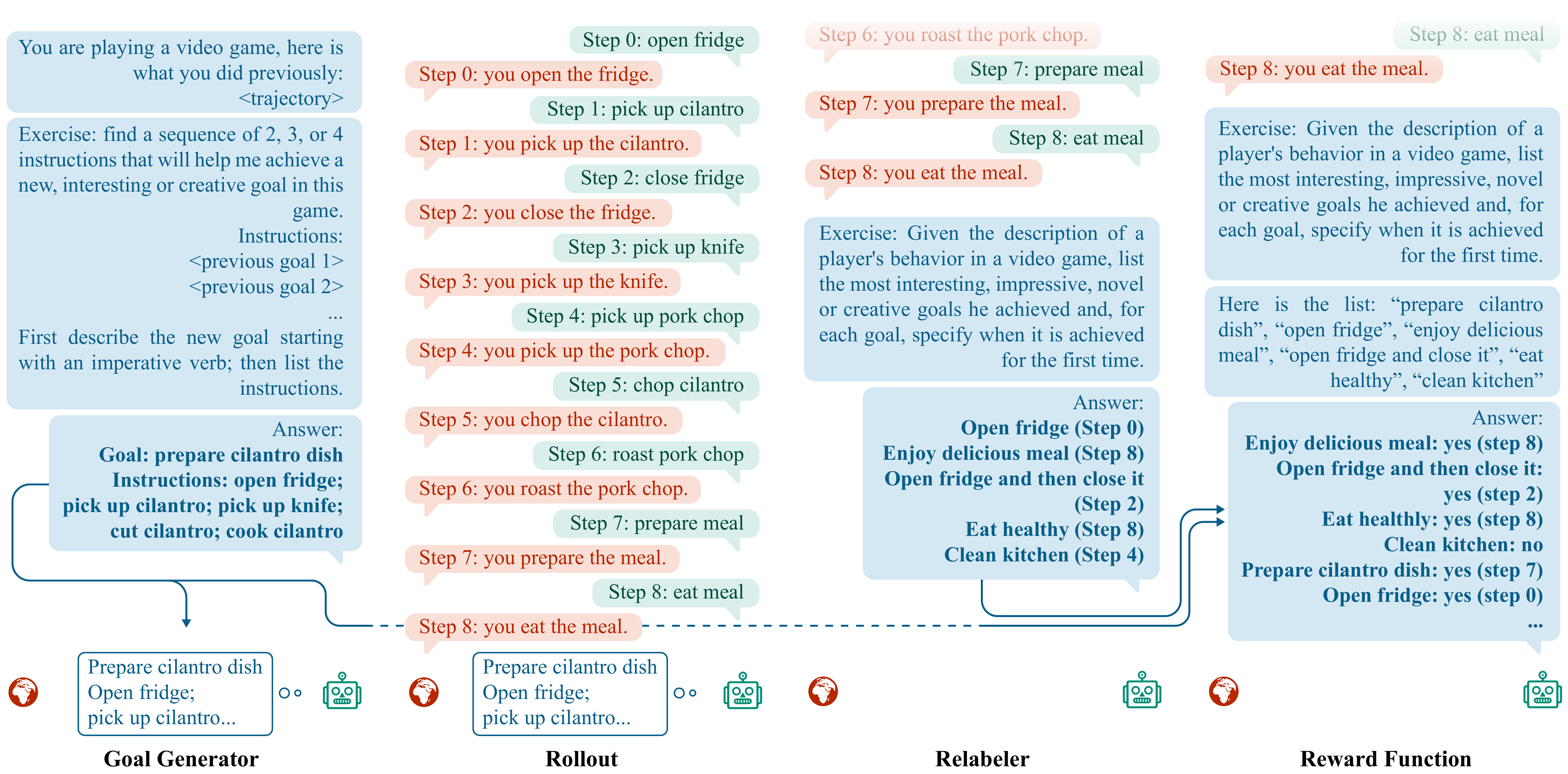}
    \caption{\textbf{Components of the Language Model Augmented Autotelic Agent (LMA3).} LMA3 agents evolve in a task-agnostic interactive text-based environment. Messages have different color depending on their source: environment is red, LM is blue, learned policy is green. \textbf{Goal Generator}: the agent prompts the LM with a previous trajectory and a list of mastered goals to generate a high-level goal and its subgoal decomposition. \textbf{Rollout}: The agent then attempts to execute the sequence of subgoals in the environment using its learned policy (green). \textbf{Relabeler}: the agent prompts the LM with the trajectory obtained during the rollout and asks for a list of re-descriptions of that trajectory (achieved goals). \textbf{Reward Function}: the agent prompts the LM with the trajectory and a list of goals to measure rewards for: the main goal, the subgoals and the goal redescription generated by the relabeler. See complete prompts in Appendix~\ref{app:prompts}.}
    \label{fig:example}
    \vspace{-1em}
\end{figure}

Human skill learning is a cultural process. Most of the goals we care about are influenced by the goals of others: we clap hands to give encouragements, we strive to finish college, or learn to play the piano. For this reason, we cannot expect autotelic agents learning in isolation to learn to represent goals that we care about, nor to bootstrap an open-ended skill learning process on their own. Just like humans, they should benefit from forms of cultural transmissions; they should learn to pursue other's goals, modify and combine them to build their own, and perhaps influence the goals of others. 

A significant part of this cultural transmission is supported by language. Transmission can be explicit when it leverages the communicative functions of language: we learn and teach via direct advice, books and online tutorials. It can also be implicit when it relies on the cognitive functions of language: words help us represent categories \citep{waxman1995words}, analogies help us represent abstract knowledge \citep{gentner2016language,dove2018language}, and linguistic productivity helps us generate new ideas by recombining known ones \citep{chomsky2002syntactic}. These cultural transmissions let us leverage the skills and knowledge of others across space and time, a phenomenon known as the \textit{cultural ratchet} \citep{tennie2009ratcheting}. 

Building on these insights, we propose to augment artificial agents with a primitive form of cultural transmission. As current algorithms remain too sample inefficient to interact with humans in real time, we leverage a pretrained language model (LM) as a (crude) model of human interests, biases and common-sense. Our proposed Language Model Augmented Autotelic Agent (LMA3) uses an LM to implement: 1)~a relabeler that describes the goals achieved in the agent's trajectories, 2)~a goal generator that suggests new high-level goals along with their decomposition into subgoals the agent already masters, and 3)~reward functions for each of these goals. We demonstrates the capabilities of this agent in a text-based environment where goals, observations and actions are all textual. Figure~\ref{fig:example} depicts the overall process. 

The relabeling functions makes use of LM's ability to segment unstructured sequences into meaningful events and to focus on the most relevant ones \citep{michelmann2023large}. This allows the autotelic agent to represent possible goals and implement a form of linguistic data augmentation \citep{xiao2022robotic}. The goal generator builds on the set of goals already achieved to suggest more abstract goals expressed as sequences of subgoals. It implements a goal-based exploration similar to the one of Go-Explore \citep{ecoffet2021first}. Finally, the reward function ensures that the agent can compute goal-completion signals, the necessary reward signals to implement any kind of goal-directed learning algorithm. Augmented by LMs, our autotelic agents learn large repertoires of skills in a task-agnostic interactive text environment \textit{without any hand-coded goals or goal-specific reward functions.}

The present paper focuses on the generation of diverse, abstract and human-relevant goals in a task-agnostic environment. Because the problem of training a policy that performs and generalizes well to a large set of goals is orthogonal to the tackled issue, we limit ourselves to a crude goal-directed learning algorithm in order to limit the computational budget of calling for an LM API (see computational budget calculations in Section~\ref{sec:discussion}). In Section~\ref{sec:discussion}, we discuss how the insights gained in this paper can be leveraged to implement more open-ended learning systems.

\section{Related Work}  

Skill learning can be modeled mathematically as a reinforcement learning problem (RL) \citep{sutton1998introduction}. In an RL problem, the learning agent perceives the state of the world $s\in\mathcal{S}$, and act on it through actions $a\in\mathcal{A}$. These actions change the world according to a stochastic dynamic function that characterizes the environment $\mathcal{T}:\mathcal{S}\times\mathcal{A}\to\mathcal{S}$. An agent learns a skill by training a policy $\Pi:\mathcal{S}\to\mathcal{A}$ to sample action sequences that maximize its expected future returns $\mathcal{R}$ computed from a predefined reward function $R:\mathcal{S}\times\mathcal{A}\times\mathcal{S}\to\mathbb{R}$ using a temporal discount factor $\gamma\in[0, 1]$ such that $\mathcal{R}=\sum_t r_t\cdot\gamma^t$. Multi-goal RL problems extend the RL problem to support the learning of multiple skills in parallel \citep{schaul2015universal}. The agent now pursues goals $g\in\mathcal{G}$, each associated with their own reward function $R_g$ and trains a goal-conditioned policy $\Pi^\mathcal{G}:\mathcal{S}\times\mathcal{G}\to\mathcal{A}$ to learn the corresponding skills. 
In RL literature, a trajectory $\tau$ is a sequence of transitions where each transition is a tuple containing information at a certain time step $t$: the state $s_t$ and the action taken $a_t$.

But where do goals come from? Most approaches hand-define the set of admissible goals and corresponding reward functions. They let agents either sample goals uniformly \citep{schaul2015universal}, or build their own curriculum \citep{portelas2020automatic}. When the goal space is large enough, this can lead to the emergence of diverse and complex skills \citep{team2023human}. Truly open-ended skill learning, however, requires the frequent update of goal representations as a function of the agent's current capabilities\,---\,only then can it be never-ending. To this end, the autotelic framework proposes to endow learning agents with an intrinsic motivation to represent and generate their own goals \citep{colas2022autotelic}.

Learning to represent, imagine and sample goals to learn skills that humans care about requires interactions with human socio-cultural worlds (see argument in introduction, and \cite{colas2022language}). Autotelic agents must first internalize the goal representations of humans before they can learn corresponding skills, build upon them and contribute back to a shared human-machine cultural evolution. Goal representations can be learned by inferring reward functions from human demonstrations \citep{ng2000algorithms, arora2021survey}, via unsupervised representation learning mechanisms \citep{warde2018unsupervised, eysenbach2018diversity, pong2019skew}, or by learning to identify the goals achieved in past trajectories from human descriptions (trajectory relabeling, \cite{andrychowicz2017hindsight, lynch2020language, xiao2022robotic}). 

Building on goal representations learned from linguistic descriptions generated by a simulated social partner, the \textit{Imagine} agent invents new linguistic goals recomposed from known ones \citep{colas2020language}. Although crude, this goal imagination system allows the agent to pursue and autonomously train on creative goals it imagines, which results in improved systematic generalization and more structured exploration. The present paper extends the \textit{Imagine} approach by leveraging powerful language models to implement several components of the autotelic agent: goal representations, goal-directed reward function and relabeling system. \textit{Imagine} required a (simulated) human in the loop to bootstrap goal representations and could only imagine slight variations of training goals due to its limited imagination algorithm, its lack of grounding and the limited generalization of its reward function. On the other hand, LMA3 does not require any human or engineer input and can generate and master a much wider diversity of goals thanks to the common-sense knowledge and generalization capabilities of LMs.

We evaluate our proposed agent in an interactive text-based environment where observations and actions are all textual \citep{cote2019textworld}. Text-based environments set aside the challenges of learning from low-level sensors and actuators and focus on higher-level issues: learning in partially observable worlds, learning temporally-extended behaviors, learning the human-like common-sense required to solve these tasks efficiently, etc \citep{he2015deep, narasimhan2015language, cote2019textworld}. Text-based environments circumvent the necessity to ground LMs into low-level sensorimotor streams (\eg visual inputs and low-level motor outputs) and let us focus on the artificial generation of more abstract, human-relevant goals. This paper is the first to implement autotelic agents with no prior goal representations in text worlds. 

Pretrained language models have recently been used to augment RL agents in various ways. In robotics setups, they were used to decompose predefined high-level tasks into sequences of simpler subgoals \citep{yao2020keep,huang2022language,huang2022inner,ahn2022can}. To limit the hallucination of implausible plans, several extensions further constrain the model by either careful prompting \citep{singh2022progprompt}, by asking the model to generate code-based policies that automatically checks for preconditions before applying actions \citep{liang2022code} or by implementing further control models to detect plan failures and prompt the LM to suggest corrected plans \citep{wang2023describe}. LMs can be used to implement a reasoning module to facilitate the resolution of sensorimotor tasks \citep{dasgupta2023collaborating}. They can be finetuned to implement the policy directly \citep{carta2023grounding}. Closer to our work, the MineDojo approach finetunes a multimodal model to implement a reward function in Minecraft and asks an LM to generate plausible goals to measure the generalization of the reward function \citep{fan2022minedojo}. Finally, the ELLM algorithm prompts an LM to suggest exploratory goals to drive the pretraining of artificial agents in Crafter and HouseKeep \citep{du2023guiding}. 

Our proposal differs from these approaches in several ways. Our agent is autotelic: it generates its own goals, computes its own rewards. In text-based games, our architecture can handle a large diversity of goals including time-extended ones that can only be evaluated over long trajectories (\eg \textit{bring the onion to the counter after you've opened and closed the dishwasher}). In contrast, the MineDojo agent has no control over its goals and is limited to generate rewards for a low diversity of short-term goals due to the limited generalization capabilities of the CLIP-based reward \citep{fan2022minedojo}. ELLM generates its own exploration goals but only considers goals that can be reported from a single state (time-specific goals). It computes rewards using the similarity between the LM-generated goal and descriptions from a captioner, which fundamentally limits the diversity of goals that can be targeted to the list of behaviors the captioner can describe. In the current setup, training or learning a captioner requires some information about the set of behaviors the agent could achieve, which limits the potential for open-ended learning. Compared to ELLM, LMA3 further endows the agent with the ability to perform \textit{hindsight learning} by relabelling past trajectories \citep{andrychowicz2017hindsight} and the ability to chain subgoals to formulate more complex goals. In contrast with previous approaches \citep{yao2020keep,huang2022language,huang2022inner,ahn2022can}, we do not leverage expert knowledge to restrict the set of subgoals but learn them online and add the possibility for the agent to use these composed goals as subgoals for future goal compositions.

\section{Methods}
\label{sec:methods}

This section introduces our learning environment and assumptions (Section~\ref{sec:settings}), as well as the proposed \textbf{Language Model Augmented Autotelic Agent} (LMA3, Section~\ref{sec:lma3}).

\subsection{Setting and Assumptions}
\label{sec:settings}

\textbf{Problem setting.}
In a task-agnostic environment, we aim to implement the automatic generation of context-sensitive, human-relevant, diverse and creative goal representations. Goal representations are not only goal descriptions but also associated reward functions. Given a sufficiently effective learning algorithm, an autotelic agent endowed with such a goal generation system should learn a large diversity of skills in task-agnostic environments.

\textbf{Learning environment.}
We place the LMA3 agent in a text-based environment called \textit{CookingWorld} \citep{cote2019textworld, madotto2020exploration}. The agent receives textual observations and acts via textual commands. Is it not provided with a predefined list of goals or reward functions. The agent is placed in a kitchen filled with furniture (7 including dining chair, fridge, counter, etc), tools (4 including knife, toaster) and ingredients (7 including potatoes, apples, parsley). Across 25 consecutive timesteps, the agent can pick up and put down objects, open and close containers, cut and cook ingredients in various ways, and finally combine them to make recipes. At any step, the agent uses its learned policy to choose an action from the \textit{admissible actions}: the subset (N\,$\approx$\,30-50) of all possible actions (N\,=\,143) that the agent can take in the current context (\eg the agent needs to find and pick up the knife before it can cut any ingredient). Examples of goals the agent could imagine and learn to master include: \textit{slice a yellow potato}, \textit{cook two red ingredients}, \textit{tidy up the kitchen by putting the knife in the cutlery drawer}, \textit{aim to use all three types of potatoes in the dish}, etc.

\textbf{Assumptions.}
We make the following assumptions: 1)~we only consider text-based environments to allow straightforward compatibility with the LM (see discussion in Section~\ref{sec:discussion}); 2)~we assume access to a language model sufficiently large to capture aspects of human common-sense and interests and allow in-context few-shot learning (see implementation aspects below); 3)~the agent is spawned in the same deterministic environment at the beginning of each episode. We use a deterministic environment for two reasons. First, because the goal generator needs to know about the environment to generate feasible goals. This is achieved by prompting the goal generator with a past trajectory in that same environment. Second, because it allows us to implement skill learning with a simple evolutionary algorithm, which considerably reduces the sample complexity and thus the cost of querying the LM\,---\,albeit to the detriment of generalization. Note, however, that robustness and generalization of acquired skills are \textit{not} the focus of this paper, which is interested in the automatic generation of diverse and human-relevant goals. In contrast to most goal-conditioned approaches, we do not assume access to a predefined set of goal representations or reward functions.

\subsection{Language Model Augmented Autotelic Agent (LMA3)}
\label{sec:lma3}

\begin{figure}[h]
    \centering
    \includegraphics[width=450pt]{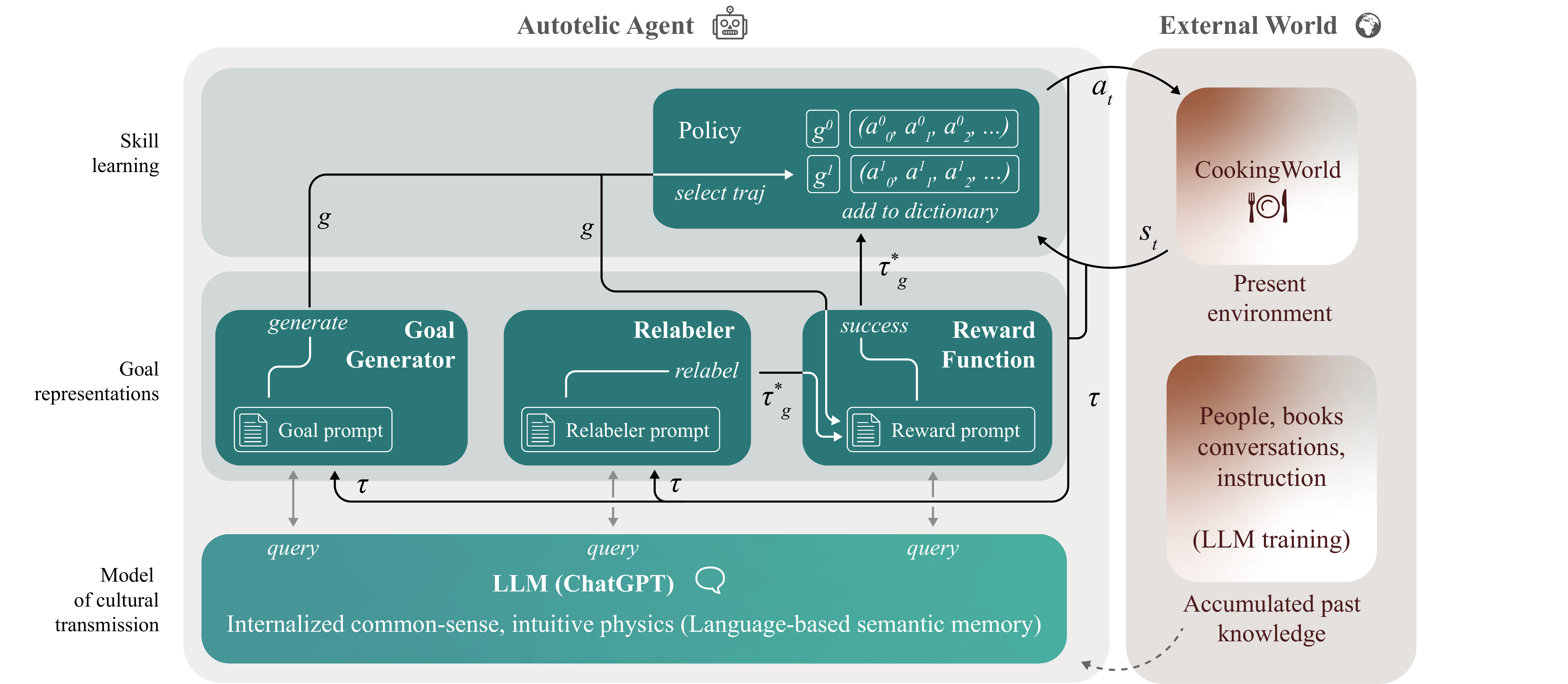}
    \caption{\textbf{General architecture of LMA3}. LMA3 assumes access to a model of cultural transmission implemented via ChatGPT (dashed line). As shown in the goal representations block, LMA3 leverages that model to generate goals $g$ (left), relabel past trajectories $\tau = (s_0, a_0, ..., s_T, a_T)$ as $\tau^*_g$ (middle) and compute rewards when goals are reached (right). 
    $s_t$ and $a_t$ denote the state representation and the action taken at game step $t$.
    The goal-conditioned policy (top) attempts to reach its goal within \textit{CookingWorld} (right) and uses relabels and rewards for learning.}
    \label{fig:architecture}
    \vspace{-1em}
\end{figure}

\textbf{General architecture.}
LMA3 augments a traditional multi-goal learning architecture with goal representations and goal generation powered by an LM (Figure~\ref{fig:architecture}). In contrast with a standard multi-goal architecture which predefines a bounded distribution of reward functions, the set of goals and associated reward functions supported by the LM is virtually unbounded; it includes all goals expressible with language. The goal generator samples a goal for the goal-conditioned policy to pursue (see \textbf{LM Goal Generator} below). Following the hindsight experience replay mechanism \citep{andrychowicz2017hindsight}, we \textit{relabel} trajectories obtained from rolling out the policy and add those to a replay buffer. The relabeling process is implemented by another LM instance which labels up to 10 goals achieved during the trajectory along with their precise completion timestamps (see \textbf{LM Relabeler} below). The goal-conditioned policy is then trained with a simple evolutionary algorithm (see \textbf{Skill Learning} below). For all prompts, we provide two examples (few-shot prompting \cite{brown2020language}) and reasoning descriptions (chain-of-thought prompting \cite{wei2022chain,kojima2022large}). We provide the complete prompts in Appendix~\ref{app:prompts}.

\textbf{LM Goal Generator.}
In the first episode, the agent does not have any goal representation and simply samples random actions. Then, the agent enters a bootstrapping phase of 4000 episodes for which it uniformly samples goals from the set of discovered goals generated by the relabeler and validated by the reward function (see \textbf{LM Relabeler} and \textbf{LM Reward Function} below). This allows the agent to first focus on simple goals. After that bootstrapping phase, the agent starts using the LM Goal Generator. At the beginning of each episode, we provide context to the LM by prompting it with the agent's trajectory in the previous episode and a list of up to 60 goals previously reached by the agent. Then, we ask it to generate a high-level goal and its decomposition in a sequence of 2--4 subgoals from the list. The decomposition lets an agent explore its environment by chaining sub-skills it knows about. This can be seen as an extension of the \textit{Go-Explore} strategy where, after achieving the first goal (\textit{go}), the exploration is further structured towards another goal (\textit{explore}) that is a \textit{plausibly useful} continuation of the first \citep{ecoffet2021first}. 

\textbf{LM Relabeler.}
After each episode, we prompt the LM with the trajectory of actions and observations and ask for a list of up to 10 descriptions of the goals achieved in the trajectory, as well as specific timestamps for when they were achieved. Note that the goal pursued by the agent in the episode does not matter, the LM Relabeler is free to provide any description of the trajectory. For each goal description we generate a positive trajectory like so: we create a sub-trajectory from step 0 to the step of goal completion, assign a positive reward to the last step and add it to the replay buffer.

We further investigate the impact of leveraging human advice nudging the agent to focus on more abstract and creative goal descriptions. We do so by replacing the 11 simple examples provided in the prompt with 11 more elaborate ones. Instead of describing simple goals involving one action on one specific object (\eg \textit{roast a white onion}), we provide examples involving sequences of actions (\eg \textit{use the oven for the second time}), conjunctions of several actions (\eg \textit{roast an onion and a bell pepper and fry carrots}) or more abstract verbs (\eg \textit{find out whether the keyholder has something on it}). In contrast with hard-coded relabeling systems, the LM Relabeler uses more linguistic diversity (\eg synonyms), can describe combinations of actions (\eg \textit{cook two onions}), or more abstract actions (\eg \textit{hide an object}).  

\textbf{LM Reward Function.}
After each episode, we prompt the LM with the trajectory and ask whether the agent achieved any of the following goals: 1)~the main high-level goal given by the LM Goal Generator, 2)~each of the subgoals (after the bootstrapping phase), 3)~each of the relabels generated by the LM Relabeler. 1 and 2 provide feedback to the agent about the goals it was attempting to reach while 3 provides a double check of the redescriptions offered by the LM Relabeler, which could be prone to hallucinations. 

\textbf{Skill Learning.}
This paper focuses on the generation of diverse, human-relevant goals and the study of a self-bootstrapped goal imagination and redescription system powered by LMs. To simplify skill learning, we consider a deterministic environment and evolve sequences of actions conditioned on the goal given a simple evolutionary algorithm. For each goal description generated by the LM Relabeler, the agent stores the sequence of actions that led to the completion of that goal in a dictionary. If this goal was achieved previously, it only stores the shortest action sequence. When prompted to achieve a sequence of subgoals, the agent chains the corresponding action sequences together to form the goal-directed policy.  Exploration is supported by two mechanisms: 1)~by chaining action sequences prompted by the LM Goal Generator and 2)~by truncating the action sequence towards the last subgoals in the chain at a uniformly sampled time with probability $\epsilon=0.2$. After executing the action sequences for all subgoals, and perhaps having truncated the last one, the agent samples actions randomly from the set of admissible actions in proportion of their rarity (\ie 1 over their occurrence).

\section{Experiments}
\label{sec:experiments}

We compare LMA3 to three ablations and an oracle baseline in the task-agnostic \textit{CookingWorld} environment to assess its learning abilities. 
For all experiments, we plot the average and standard deviations across 5 seeds. For the different LM modules, we use ChatGPT (\textit{gpt-3.5-turbo-0301}) as provided through the OpenAI API \citep{brown2020language}. The code will be released publicly with the camera-ready version of the paper.

\textbf{Ablations and oracle baseline.}
LMA3 is the first algorithm to allow the automatic generation of linguistic goals and reward functions with no interventions from the engineer. For this reason, there were no obvious baselines to compare LMA3 with. We consider three ablations that remove: 1)~the use of human advice in the prompting of the LM Relabeler (\textit{LMA3 $\backslash$ Human Tips}), 2)~the use of human advice and the LM Goal Generator (\textit{LMA3 $\backslash$ LM Goal \& Human Tips}), 3)~the use of human advice and chain-of-thought prompting  (\textit{LMA3 $\backslash$ CoT \& Human Tips}). In the absence of the LM Goal Generator, goals are uniformly sampled from the set of goals previously discovered, just like in the bootstrapping phase of the LM Goal Generator (see Section~\ref{sec:methods}). Our baseline is a standard goal-conditioned agent trained on a hard-coded set of 69 goals involving picking up, cooking or cutting objects in the text-based environment (\textit{Hardcoded Oracle Baseline}). This baseline samples goal uniformly from the set of goals previously discovered and uses an oracle relabeler. It implements a standard goal-conditioned policy learning algorithm in text-based environments.

\textbf{Performance on a human-defined goal space.}
We want to measure the ability of autotelic agents to learn skills that humans care about. However, autotelic agents learn their own goal representations in worlds that can afford a large space of possible actions and, for this reason, there is no objective set of goals these agents should learn about, no objective evaluation set. We hand-defined a set of 69 evaluation goals involving picking up, cooking or cutting objects in the text-based environment (see list in Appendix~\ref{app:evalset}). This list is obtained by applying each of the possible action types of the agent (\eg slice, dice, roast, pick up, put, open, close) to each possible object (\eg slice+ingredient, open+container) and adding the goal of preparing the recipe the agent can find in the cookbook present in the \textit{CookingWorld}. While the \textit{Hardcoded Oracle Baseline} is explicitly trained on this set of goals and can make use of an oracle reward function and relabeling function, LMA3 variants are not given any prior knowledge of these goals.

\begin{figure}[!t]
    \centering
    \subfigure[\label{fig:eval_perf_hardcodeda}]{\includegraphics[width=0.35\textwidth]{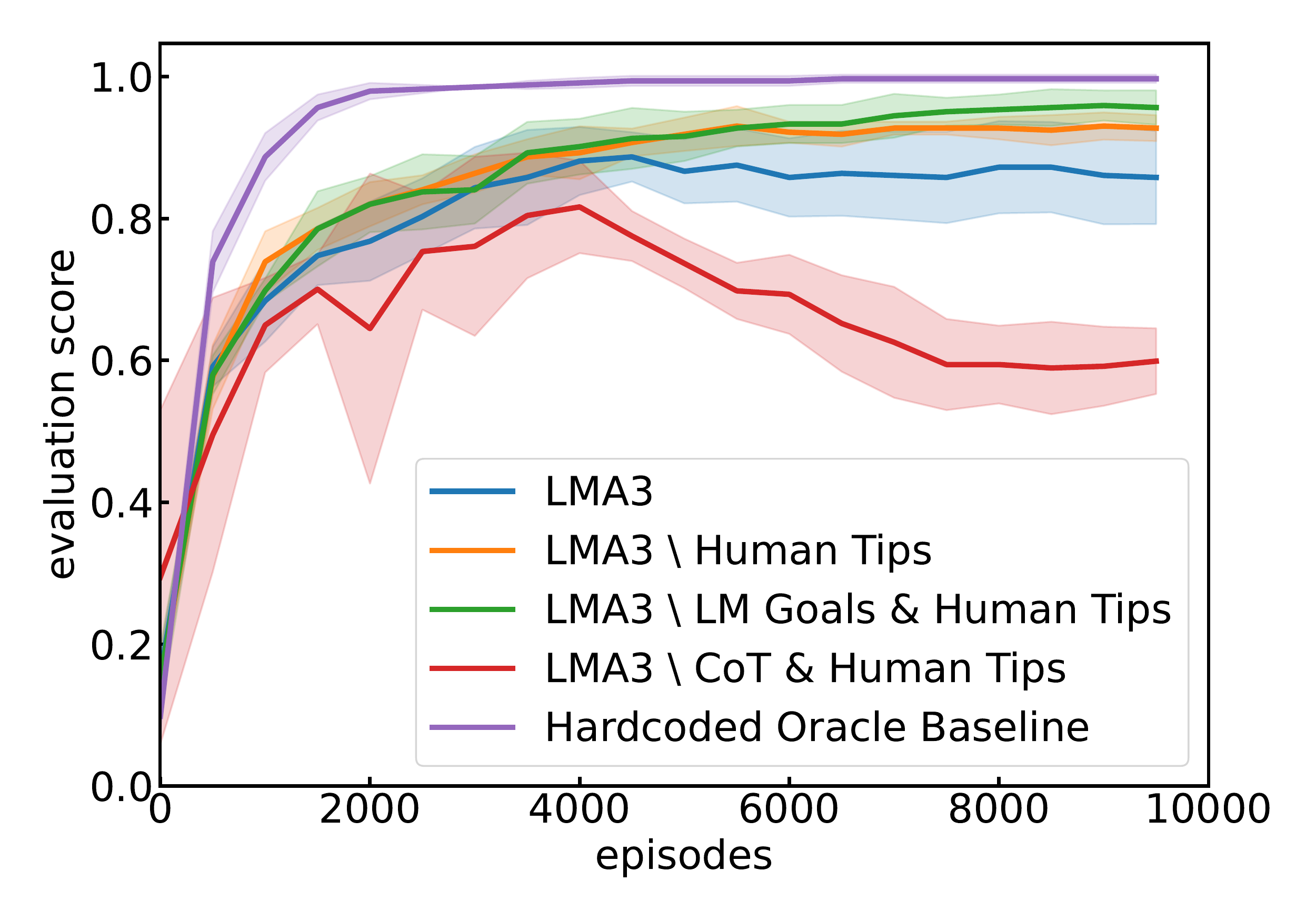}}   
    \hspace{1.5cm}
    \subfigure[\label{fig:eval_perf_hardcodedb}]{\includegraphics[width=0.4\textwidth]{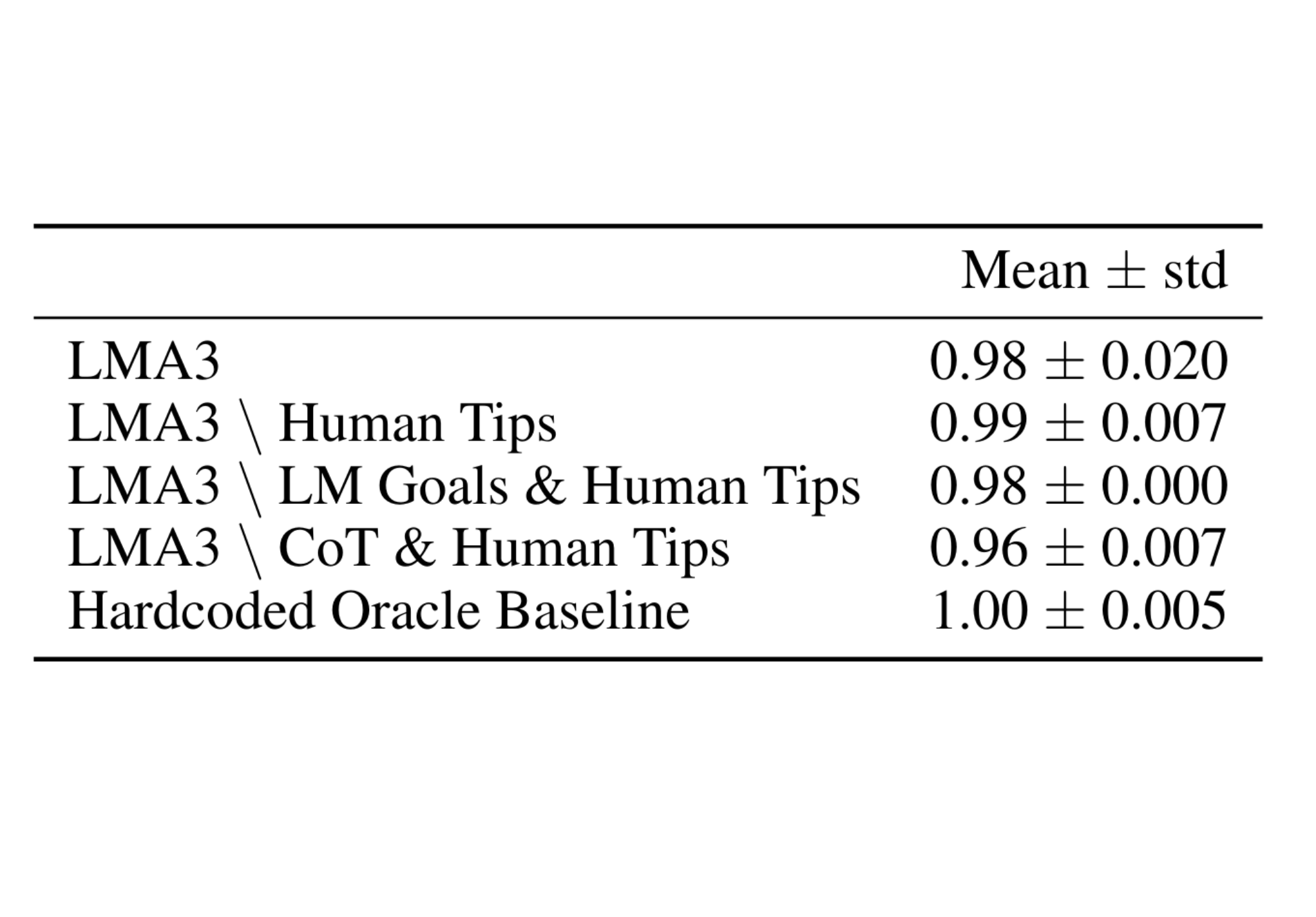}}    
    \caption{\textbf{Performance on human-defined goal space}. Performance on the hand-coded evaluation set containing 69 human-relevant goals, as measured by hard-coded reward functions. (a) across training; (b) at the end of training, after relabeling with the oracle relabeling function and \textit{without further interactions}.}
    \label{fig:eval_perf_hardcoded}
\end{figure}

Figure~\ref{fig:eval_perf_hardcoded} presents the success rates on this evaluation set computed with the hard-coded reward functions corresponding to each of the 69 goals (not given to LMA3 agents). This metrics evaluates a \textit{minimal requirement}: can LMA3 agents learn to master some of the goals human care about in this world without assuming predefined representations for these goals? Most LMA3 variants learn to reach a large fraction of the evaluation goals, which indicates that leveraging goal representations captured by LMs may support the autonomous learning of skills that humans care about. The \textit{Hardcoded Oracle Baseline} makes use of an \textit{oracle relabeling function} that can faithfully detect any of the 69 goals when it is reached in a trajectory. If we now provide this function to a trained LMA3 agent, we can sweep its memory of action sequences and find the ones achieving the evaluation goals. This form of finetuning does not require further interactions in the environment. Applying it further boosts the success rates of LMA3 agents to near perfect results (see Figure~\ref{fig:eval_perf_hardcodedb}). This shows that LMA3 agents do reach the evaluation goals but sometimes fail to relabel them properly and instead choose to focus on describing other demonstrated behaviors. After finetuning, some of the LMA3 seeds manage to complete the recipe found in the cookbook: a preparation that includes picking up cilantro and parsley from the fridge, opening the cupboard, taking the knife, slicing the parsley, preparing and eating the meal (2 \textit{LMA3} seeds and 3 \textit{LMA3 $\backslash$ Human Tips} seeds). These results confirm that LMA3 can learn human-relevant goals \textit{completely autonomously}, without relying on any predefined goal representations or reward functions. 

\textbf{Skill diversity.}
We measure the diversity of discovered skills with 1)~the raw number of distinct goals; 2)~Hill's numbers, a metric inspired from the evaluation of species diversity in ecosystems \citep{jost2006entropy} and 3)~a metric inspired from the h-index used in research. Hill's numbers allow to make a distinction between species-level diversity (here the type of action required by the goal) and individual-level diversity (\eg the object on which the action is performed; the object used to perform the action, the location, etc). The type of action required by a goal is inferred as the linguistic stem of the first word of the goal's description. The goal \textit{cutting the apple} is an individual goal of the \textit{cut} goal species. Hill's numbers provide measures of diversity computed as:
\begin{equation*}
\small
    D^q=\left(\sum_{s\in \text{stems}}p_s^q\right)^\frac{1}{1-q},
\end{equation*}
\vspace{-0.3em}
where $p_s$ is the empirical fraction of goals with stem $s$ and $q$ controls the sensitivity of $D^q$ to rare vs. abundant species: $D^0$ is the count of stems, also called species richness (no sensitivity to species abundance) and $D^1$ is the exponential of Shannon's entropy, also called perplexity, a measure that quantifies the uncertainty in predicting the stem's identity of a goal uniformly sampled from the set of discovered goals \citep{jost2006entropy}. Lower $q$ puts more emphasis on species diversity while higher $q$ puts more emphasis on individual diversity. Note that stems can hide part of the information: different stems might refer to similar behaviors (\eg \textit{grab} vs \textit{pick up}) while a same stem might refer to different behaviors (\eg \textit{pick up the apple} vs \textit{pick up the meal}; the 2nd requiring a more complex and time-extended behavior). Such analysis still provides complementary information to the simple goal count. Finally, the stem h-index is computed as the maximum value $h$ such that $h$ stems have at least $h$ goals. This metric is used in research to compute a score mixing diversity and quality of paper citations, here it is used as a way to balance species-level (action type) and individual-level (the rest) diversities. 

\begin{figure}[b!]
    \centering
    \subfigure[]{\includegraphics[width=0.245\textwidth]{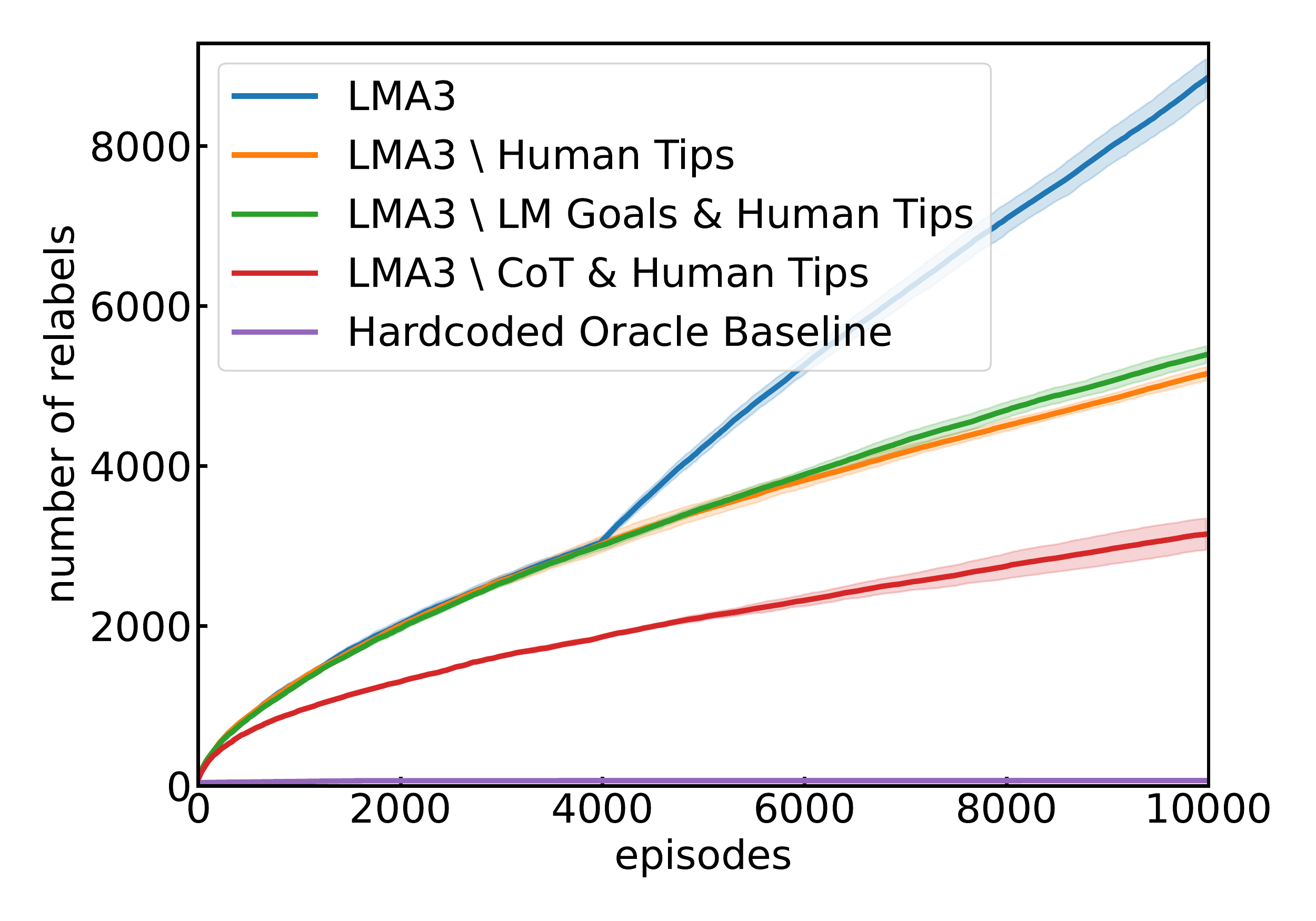}}   
    \subfigure[]{\includegraphics[width=0.245\textwidth]{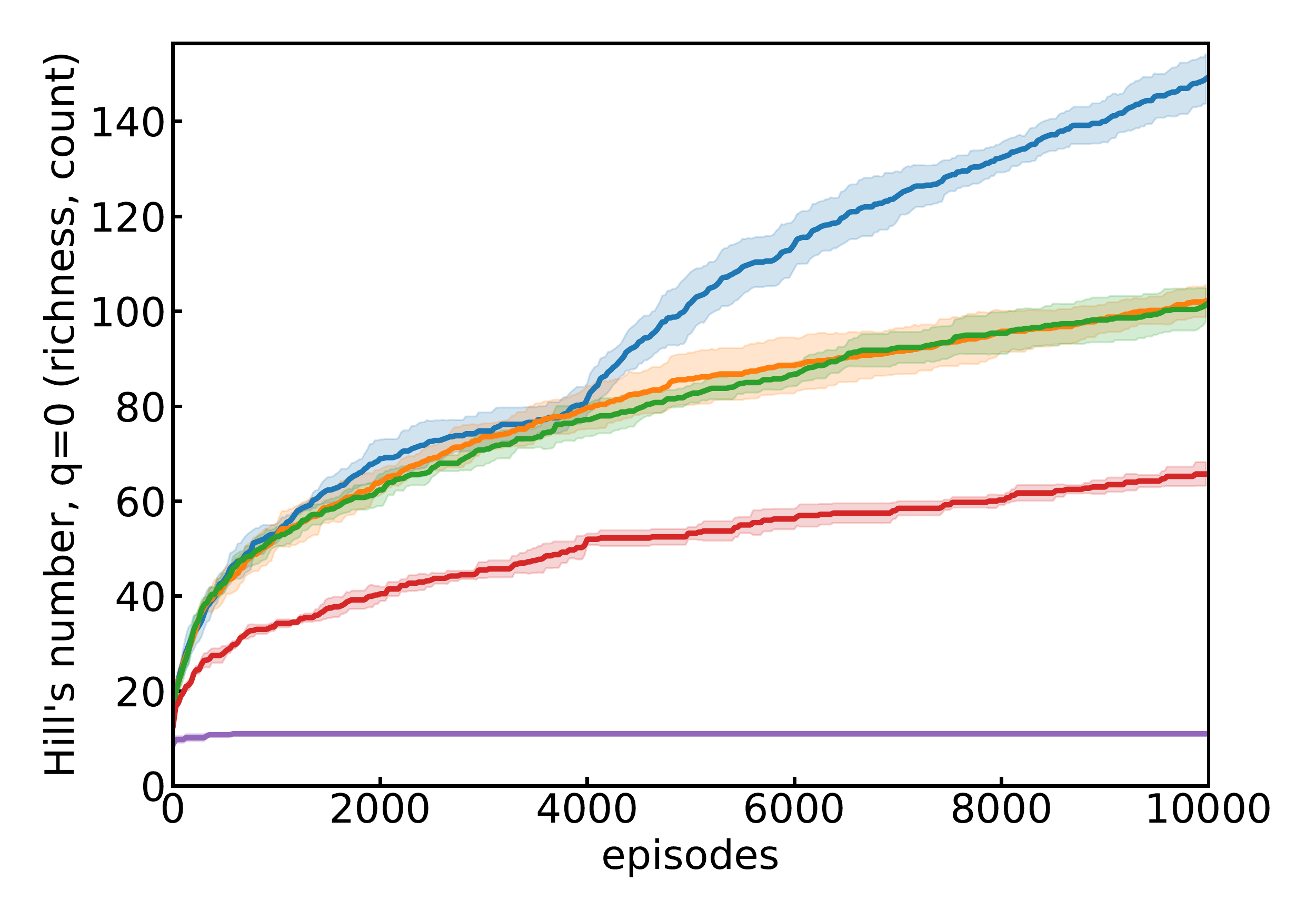}}   
    \subfigure[]{\includegraphics[width=0.245\textwidth]{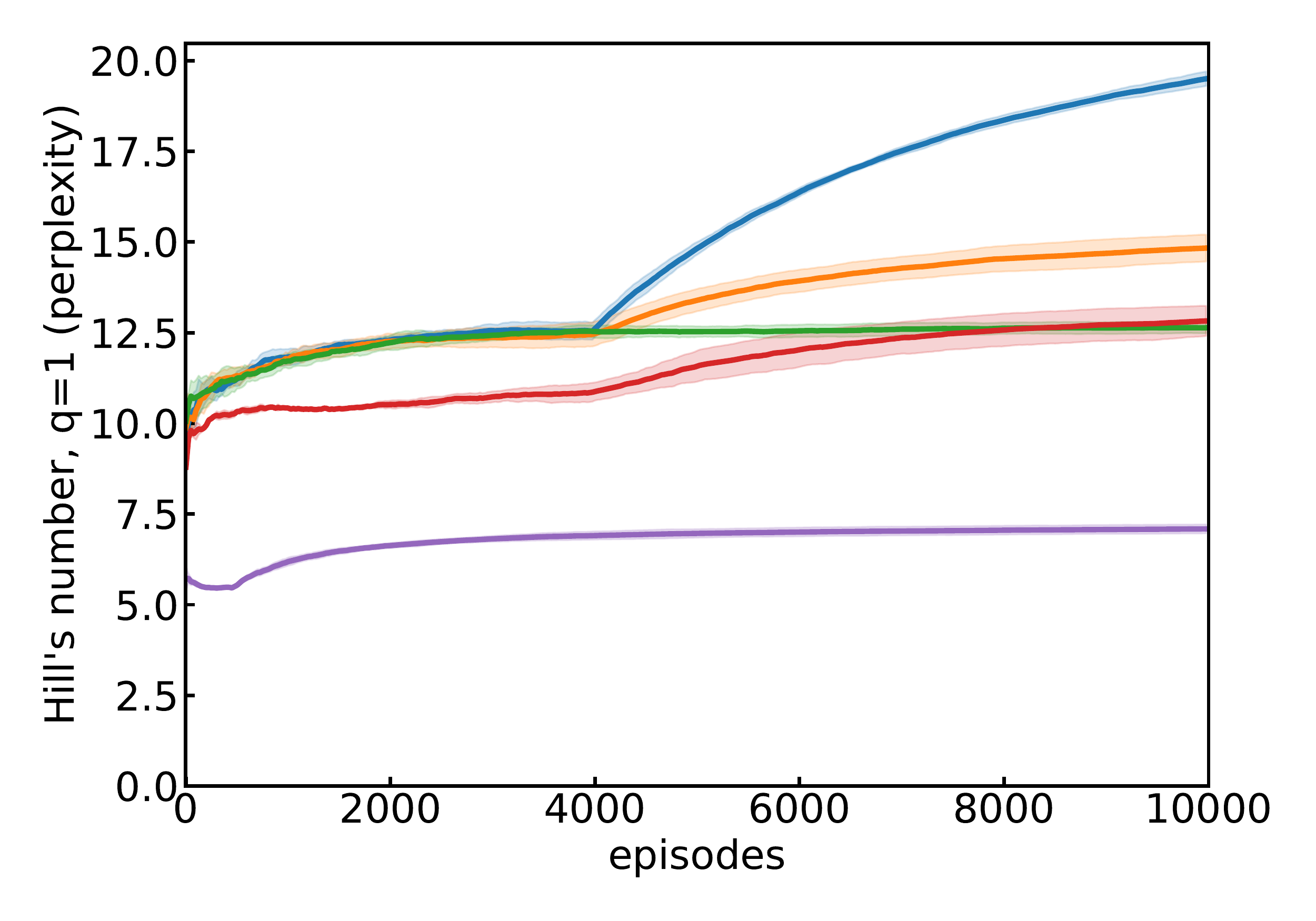}}   
    \subfigure[]{\includegraphics[width=0.245\textwidth]{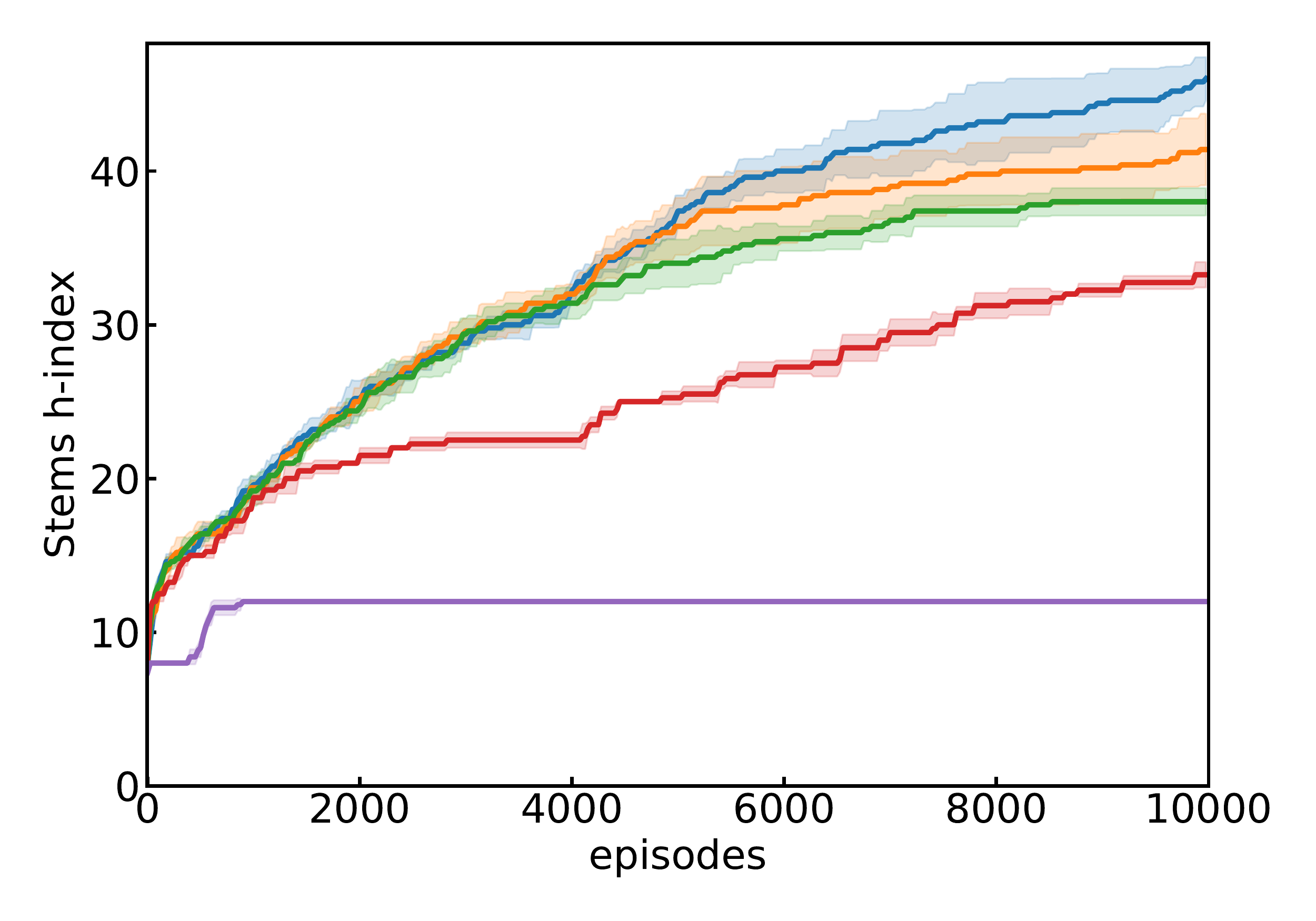}}   
    \caption{\textbf{Diversity of achieved goals}: (a) number of relabels, (b) number of stems (Hill's number with q=0), (c) perplexity (Hill's number, p=1), (d) stem's h-index. LMA3 discovers and masters a more diverse set of goals than its ablations. The \textit{Hardcoded Oracle Baseline} is limited  to discover goals from the hand-defined set of 69 goals and thus demonstrates very little diversity.}
    \label{fig:diversity}
\end{figure}

Figure~\ref{fig:diversity} reports the total number of discovered goals, Hill's numbers for $q\in\{0,\,1\}$ as well as the stem h-index. While the introduction of the LM Goal Generator (episode 4000) introduces a slight boosts in the diversity of discovered goals (orange vs green), the addition of human advice triggers a lasting increase in goal diversity (blue vs orange). Removing CoT prompting dramatically decreases the diversity of goals discovered by LMA3 (red vs orange). Finally, the diversity of the \textit{Hardcoded Oracle Baseline} remains low as it is restricted to consider the 69 goals from the hand-defined set of goals. LMA3 discovers around 9000 distinct goal redescriptions in 10000 episodes. 

\begin{figure}
    \centering
    \subfigure[]{\includegraphics[width=0.33\textwidth]{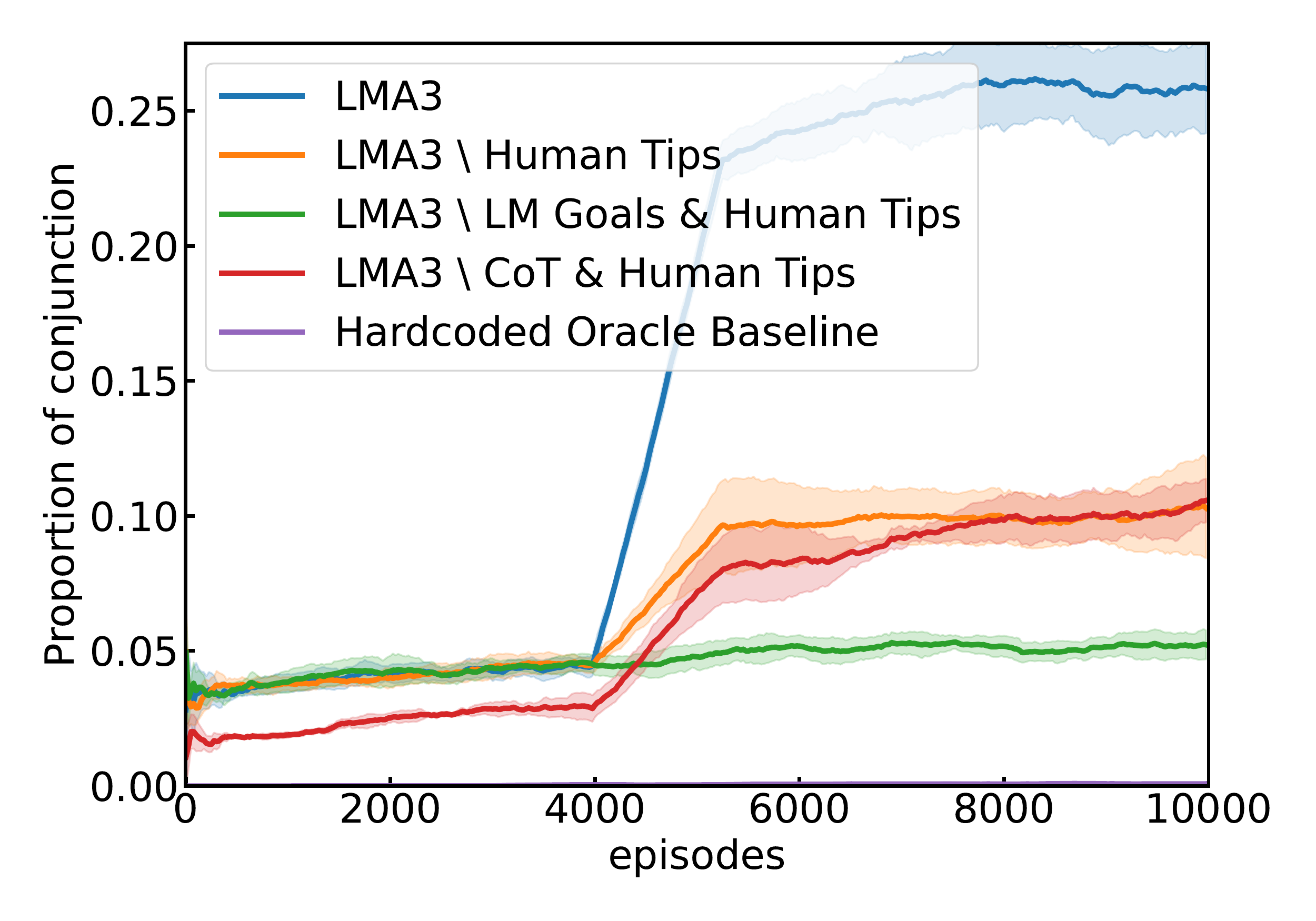}}   
    \subfigure[]{\includegraphics[width=0.33\textwidth]{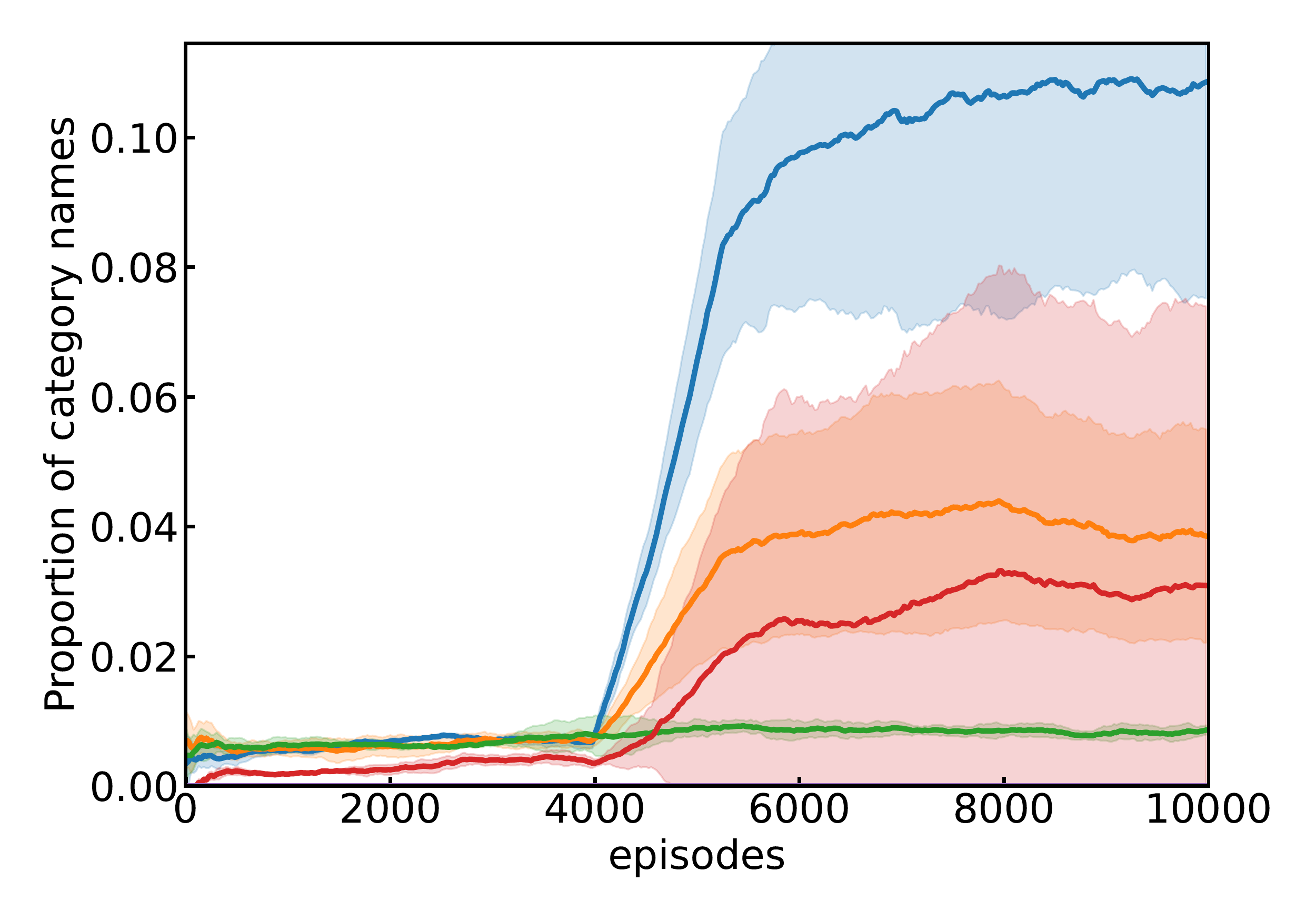}}   
    \caption{\textbf{Discovery of more complex and abstract goals}. LMA3 discovers and masters more complex goals expressed as combinations of simpler goals, or using category names (\eg \textit{ingredients}, \textit{containers}) instead of specific object name as they appear in \textit{CookingWorld} (\eg \textit{yellow potato}, \textit{kitchen drawer}).}
    \label{fig:abstraction}
    \vspace{-0.8em}
\end{figure}

\textbf{Sensitivity to human advice.}
We measure the sensitivity of LMA3 agents to a small amount of human advice. LMA3 leverages human advice in the prompt of the LM Relabeler, with only a few examples nudging the relabeler to describe more abstract behaviors involving conjunctions of actions (\eg \textit{roast an onion and a bell pepper and fry carrots}), repetition of actions (\eg \textit{open three containers}), non-specific object references (\eg \textit{slice and cook an orange ingredient}) or more abstract action predicates (\eg \textit{find out whether the keyholder has something on it}), see full prompts in Appendix~\ref{app:prompts}. Although the conditions with/without human advice use the same number of examples in their prompts, the more abstract examples used in the full LMA3 condition drives a significant increase in both the diversity of discovered goals (Figure~\ref{fig:diversity}) and in the abstraction of these goals as measured by the proportion of goals containing conjunctions (``and'', ``two'', ``three'', or ``several times''),  or category names instead of specific object names (``ingredients'', ``items'', ``container'', ``somewhere'', ``fruit'', ``vegetable'', or ``tool'') see Figure~\ref{fig:abstraction}. Note that the effect on category names, although significant, remains small. 

\begin{figure}[b]
    \vspace{-1em}
    \centering
    \subfigure[]{\includegraphics[width=0.29\textwidth]{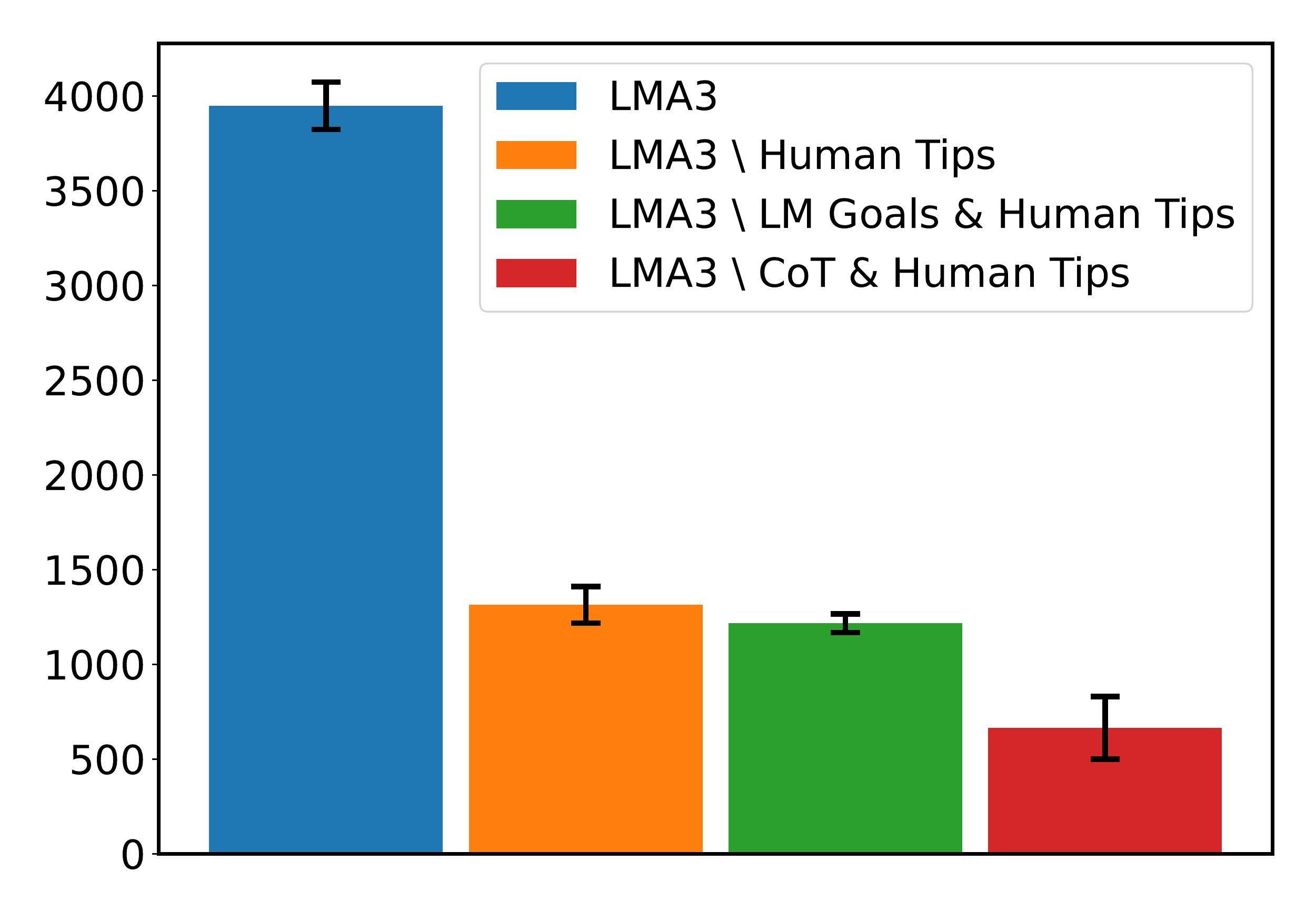}}   
    \subfigure[]{\includegraphics[width=0.29\textwidth]{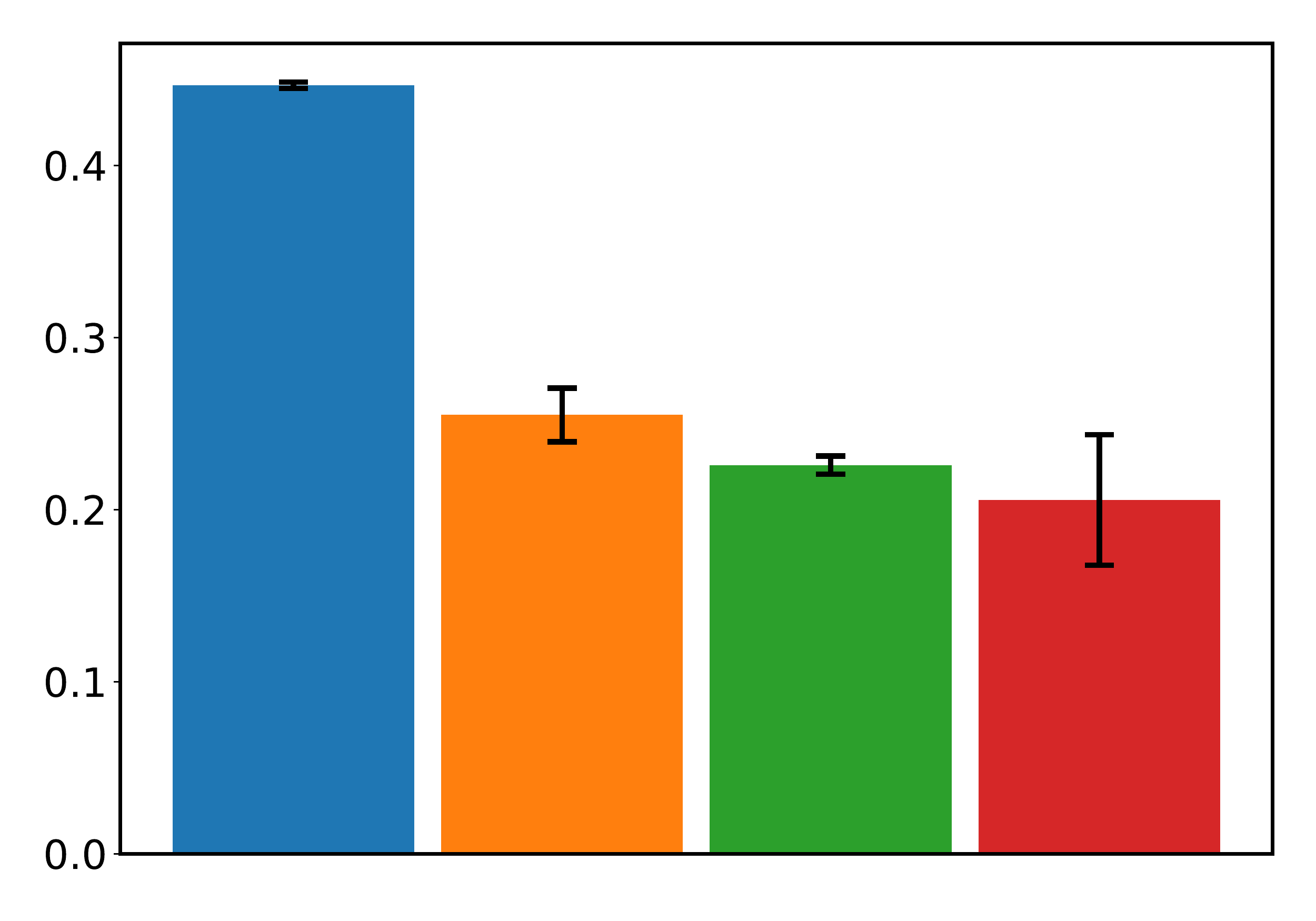}}   
    \subfigure[]{\includegraphics[width=0.29\textwidth]{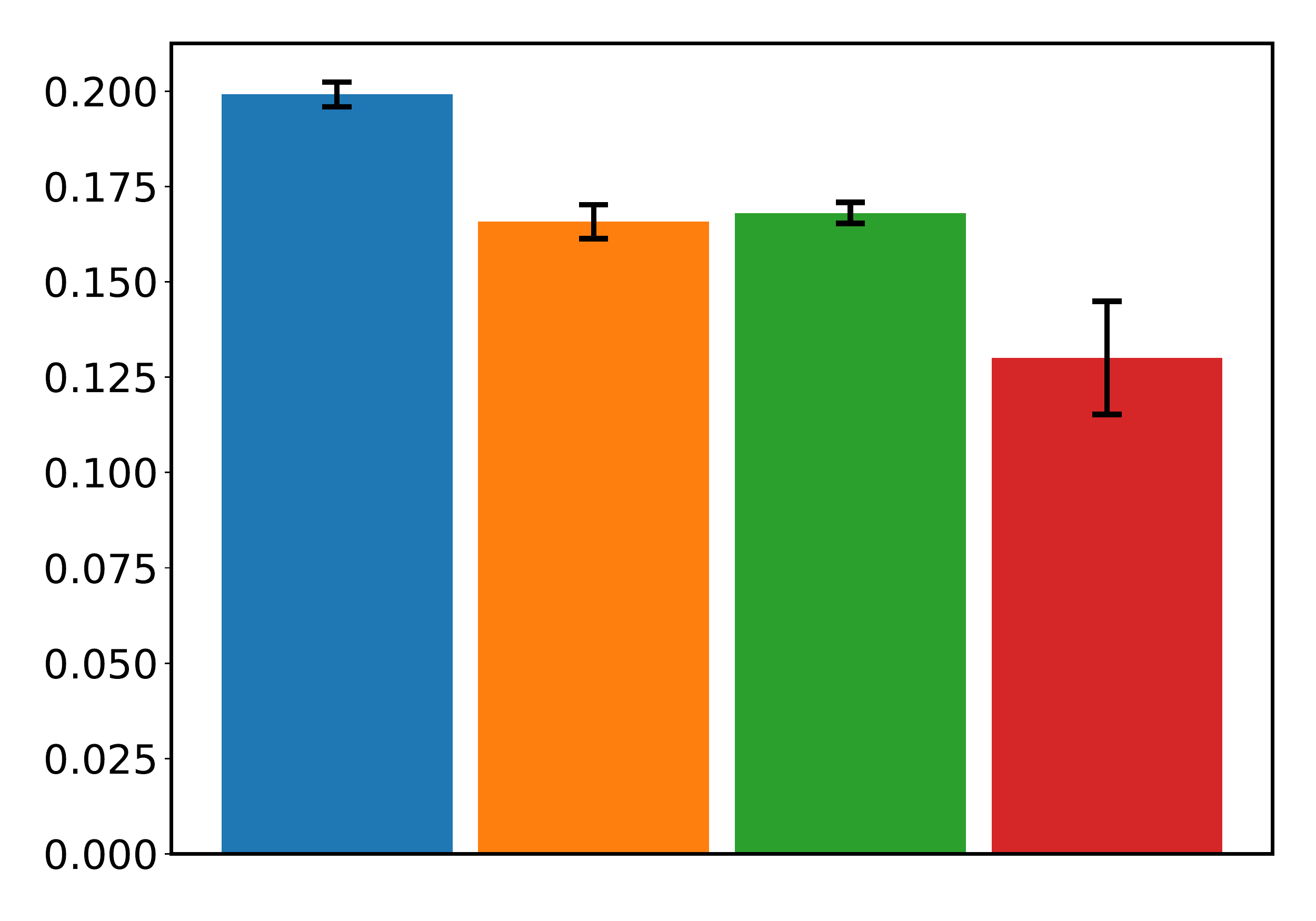}}   
    \caption{\textbf{Discovery of unique goals}. For each algorithm, average number of \textit{unique goals} that each agent was the only one to discover (a), ratio of \textit{unique goals} over the count of all goals discovered by the agent (b), and average novelty of the unique goals computed in sentence embedding space (c).}
    \label{fig:unique}
    \vspace{-0.5em}
\end{figure}

\textbf{Discovery of unique goals.}
We measure the ability of each agent to discover \textit{unique goals}; goals that were discovered by this particular seed but were not encountered by any other seed across all algorithm conditions. For each seed, we compute the number of such unique goals, report the ratio of that number over the count of all goals discovered by the agent, and give a measure of the novelty of these unique goals by reporting their average distance to the nearest neighbor in the linguistic embedding space of all goals discovered by all seeds of all conditions. This embedding is computed from a pretrained SentenceBERT model \citep{reimers2019sentence}. Figure~\ref{fig:unique} shows that LMA3 agents discover more unique goals, not only in number but also in proportion of the total number of goals they discover (b) and that these goals are more novel in average (c). 

\begin{figure}
    \vspace{-0.8em}
    \subfigure[\label{fig:selfevala}]{\includegraphics[width=0.5\textwidth]{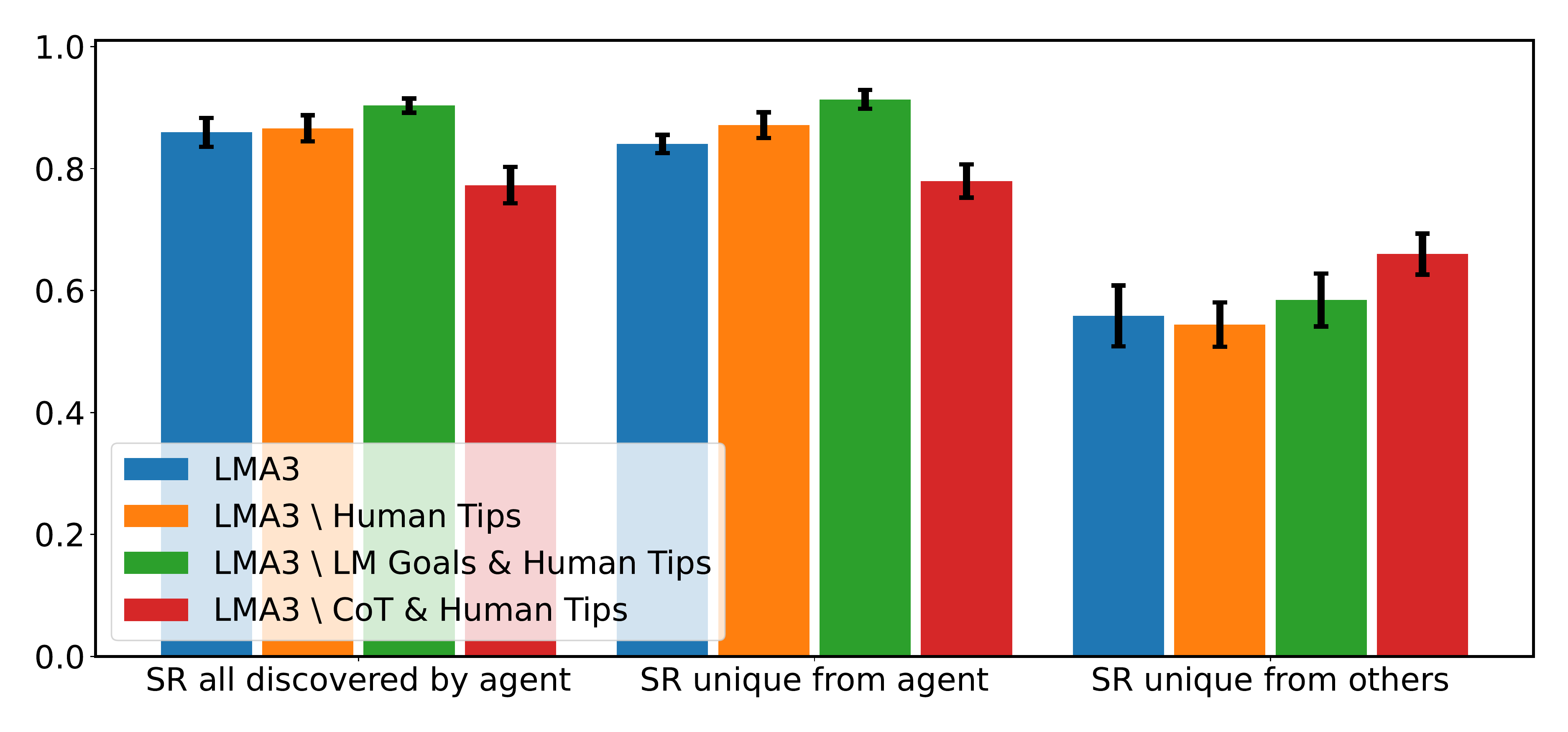}} 
    \hspace{1cm}
    \subfigure[\label{fig:selfevalb}]{\includegraphics[width=0.35\textwidth]{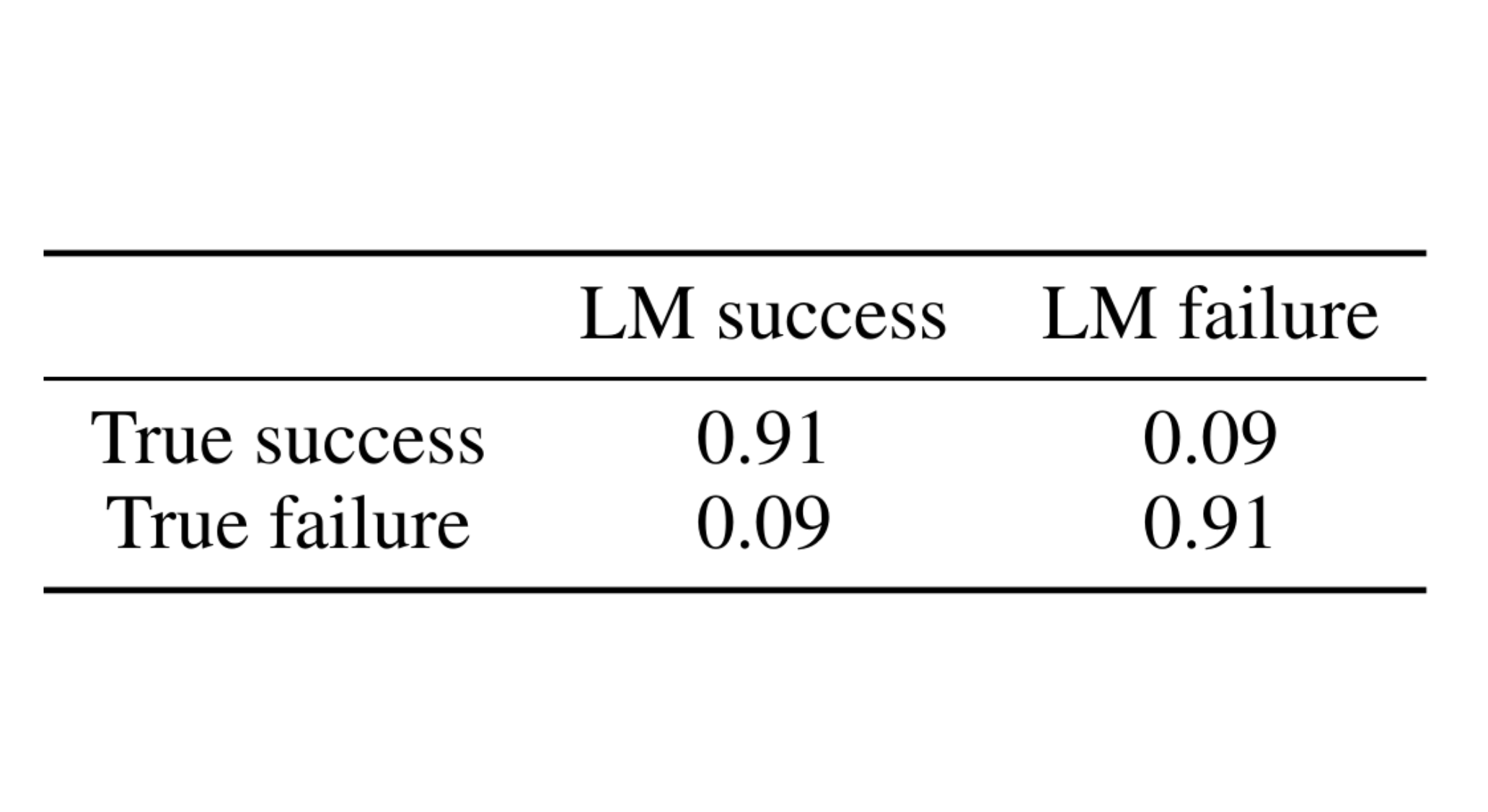}}   
    \caption{\textbf{Self-evaluated performance}. (a): Average success rates across seeds for each algorithm when computed on goals discovered during training (left), on unique goals discovered by the agent and no other agent (middle), on a sample of unique goals from other agents, not discovered by the evaluated agent (right). (b): confusion matrix of the LM Reward Function tested over 100 human-relabeled trajectories (56 true success / 44 true failures).}
    \label{fig:selfeval}
    \vspace{-0.8em}
\end{figure}

\textbf{Mastery of discovered goals.}
We measure the skill mastery of agents on several subsets of the goals they discover.  For each seed, we compute the success rate of the corresponding agent on three sets of 200 evaluation goals: 1)~a uniform sample of all goals discovered by the agent; 2)~a uniform sample of the goals only they discovered (unique goals) and 3)~a uniform sample of the set of all unique goals from other agents (Figure~\ref{fig:selfevala}). We evaluate successes with the LM Reward Function. Note that although the environment is deterministic, mistakes in the LM Relabeler or the LM Reward Function could lead agents to mistakenly classify a goal as reached when it is not. To measure the reliability of this imperfect reward function, we computed its confusion matrix given a set of N=100 trajectories using human labels as ground truth. We estimate the probabilities of both false positives and false negatives to 9\% (see confusion matrix in Figure~\ref{fig:selfevalb}).

\textbf{Examples of discovered goals.}
As discussed above, LMA3 agents discover most of the hand-defined evaluation goals (\eg \textit{slice a yellow potato}, \textit{pick up the knife}, \textit{open the fridge}). They also learn to consider goals expressed as conjunctions or disjunctions of simpler goals (\eg \textit{cook two red ingredients}, \textit{put a potato red or yellow in the kitchen cupboard}, \textit{examine an object in the kitchen, like the oven, yellow potato, green or red apple}). They sometimes express goals in more abstract ways (\eg \textit{wield the knife}, \textit{waste food}, \textit{tidy up the kitchen by putting the knife in the cutlery drawer}, \textit{pack potatoes and apples in the dishwasher}, \textit{refrigerate the yellow apple}). They can use object attributes or object categories to refer to sets of objects (\eg \textit{put a yellow ingredient in the kitchen cupboard}, \textit{aim to use all three types of potatoes in the dish}, \textit{choose to place an ingredient on a dining chair instead of the counter}). In contrast with hand-defined goals, these goals use new words, category words, or abstract action predicates that do not appear in the vocabulary of the text-based environment. 

Finally, the LM Goal Generator generates more complex goals and decomposes them into subgoals the agent masters: \eg \textit{rearrange the yellow apple and yellow potato inside the kitchen cupboard}~$\to$~[\textit{pick up the yellow apple}, \textit{take the yellow potato}, \textit{place the yellow potato in the kitchen cabinet}, \textit{place the yellow apple in the cupboard}]; \textit{assemble a meal with a fried yellow potato and a roasted red apple on a dining chair}~$\to$~[\textit{fry the yellow potato}, \textit{cook a red apple in the oven}, \textit{place the red apple and the yellow potato on the dining chair}]; \textit{serve a meal consisting of a roasted sliced yellow potato and a fried diced green apple}~$\to$~[\textit{dice and fry the green apple}, \textit{slice a yellow potato}, \textit{cook the yellow potato in the oven}]. We found that the LM Goal Generator only generated a low diversity of complex goals: they all prompt the agent to prepare some form of recipe and vary in the properties of the recipes (\eg \textit{vegetarian}, \textit{colorful}), or its particular ingredients (\eg \textit{with roasted sliced potatoes}). This behavior stems from the fact that \textit{CookingWorld} is a relatively narrow environment: a kitchen with ingredients and kitchen appliances. 

\textbf{Conceptual comparisons to other skill discovery approaches. } Implementing the \textit{Imagine} agent would require the definition of 1)~a simulated human describing some of the agent's behavior (\eg the 69 oracle goals) and 2)~a hard-coded symbolic goal imagination system \citep{colas2020language}. The original imagination system would be limited to imagine linguistic recombinations of the training goals: \eg from \textit{cut the apple}, \textit{pick the apple} and \textit{cut the parsley}, \textit{Imagine} could generate \textit{pick up the parsley}. These recombinations are either non-semantic (\eg \textit{open the parsley}) or already contained in the set of 69 goals. As a result, Imagine would be strictly less powerful than the Oracle Baseline: it would not imagine more goals but would need to learn a reward function from descriptions where the Oracle Baseline assumes oracle reward functions. 

Other skill discovery methods are limited to low-level goal representations. Visual goal-conditioned approaches learn goal representations by training a variational auto-encoder on experienced visual states and train the agent to imagine and reach new goals in that visual embedding space \citep{pong2019skew}. One could imagine a variant of these approaches embedding linguistic trajectories in such a generative model but the resulting skills would not be particularly semantically meaningful. Unsupervised skill discovery approaches co-train a skill discriminator and a skill policy with an empowerment reward \citep{eysenbach2018diversity}. A variant of these approaches for CookingWorld could consist in co-training a captioner (skill discriminator) and a policy to maximize the likelihood of the trajectory's caption being the original goal of the policy. This algorithm does not exist yet and it is unclear how the agent should sample goals.

\section{Conclusion and Discussion}
\label{sec:discussion}

This paper introduced LMA3, an autotelic agent augmented with a language model capturing key aspects of human common-sense and interests to support the generation of diverse, abstract, human-relevant goals. In the \textit{CookingWorld}, LMA3 can learn a large set of skills relevant to humans without relying on any predefined goal representations or reward functions. The diversity of goal representations is further impacted by careful prompting involving chain-of-thought reasoning \citep{kojima2022large}, a small quantity of human-generated advice and the use of an LM-based goal generator. 

LMA3 can be applied in any environment where the agent's behavior can be described with language. Although it does not cover all possible scenarios, many of the skills that humans care about can be described with language: from simple actions like picking up a glass, to more abstract behaviors like composing a haiku or coding a sorting algorithm. Lower-level behaviors, on the other hand, are hard to express with language (\eg fine-grain robotic manipulation). For such behaviors, future work could combine LMA3 with unsupervised skill discovery algorithms such as DIAYN \citep{eysenbach2018diversity} or Skew-Fit \citep{pong2019skew}. Modular autotelic agents could then target goals from several goal spaces in parallel and perform cross-modal hindsight learning as proposed in \cite{colas2019curious}.

This paper focused on goal generation and used a simple skill learning approach to limit the sample complexity of the experiments. 
Future work could build on the proposed approach to achieve more open-ended skill learning. Let us discuss the key elements that should be improved to that end. A key aspect of open-ended learning is the co-adaptation of the goal generator and the goal-reaching policy. In LMA3, goal generation evolves as the LM Goal Generator recursively composes subgoals the agent masters towards more complex goals. In addition, the LM Relabeler should also adapt and describe harder and harder goals as the agent learns to master them. To this end, the agent should be given the ability to track its own performance (estimated success rate, learning progress or uncertainty) and use these metrics as intrinsic motivations to guide relabeling. Improving the skill learning algorithm would also help LMA3 generalize to a larger diversity of goals and thus focus on harder ones faster. Examples of more sophisticated skill learning algorithms include: leveraging deep reinforcement learning approaches \citep{hessel2018rainbow} with transformer-based architectures \citep{Chen2021DecisionTR,Janner2021ReinforcementLA}, finetuning another large language model using online interactions \citep{carta2023grounding}, or leveraging state-of-the-art model-based approaches \citep{hafner2023mastering}. 

Given these more scalable learning approaches, one should consider the exploration of larger worlds. The set of possible interactions in \textit{CookingWorld} is fundamentally limited to a number of distinct interactions with a few objects. As goal generation gets more abstract and diverse and the skill learning approach learns more skills, the main bottleneck becomes the complexity of the environment. While current text-based environments are typically not open worlds, one might consider the use of non-textual open worlds such as Minecraft, coupled with image- or video-to-text captioning systems. To this day, open-source multimodal model cannot relabel trajectories in an open-ended way, and even state-of-the-art closed-source variants still require finetuning before they can be used as success detectors in specific environments \citep{du2023vision}. The multimodal version of GPT-4 may change that in the near future, and could be easily integrated within the LMA3 framework. 

Although LMs are a useful resource, they remain expensive not only to train but also to use. The experiments presented in this paper represent 550k calls to ChatGPT of about 4k tokens per call. This represents $\approx$ USD 4,400 with the public pricing of USD 0.002 /1k tokens. A given seed of LMA3 run for 10k episodes and costs about USD 240. Training large neural policies that generalize well in complex environments would require about two orders of magnitudes more episodes ($\approx$1M), which would raise the cost of any single seed to $\approx$ USD 22,400. Pushing these ideas forward may thus require a combination of reduction in inference costs and prices, the distillation of LMs into smaller environment specific reward functions and relabeling functions. 

As the field moves towards more and more open-ended agents, their evaluation becomes more complex. How should we evaluate agents that imagine their own goals, specialize in certain skills and not others? This paper used the diversity of stems as a proxy for the diversity of interaction types the agent could learn to demonstrate. Other metrics could include the measure of diversity in linguistic space, various measures of exploration computed from the agent's behavior, human studies to evaluate the diversity, creativity, abstraction and complexity of the mastered skills. The evaluation of the relevance of learned skills for humans could also require humans in the loop interactively testing the capacities of agent in standardized interaction protocols. 


\bibliography{biblio}
\bibliographystyle{collas2023_conference}

\newpage
\appendix
\section{Hand-Coded Evaluation Set}
\label{app:evalset}
Here are the 69 hand-coded goals: 
cook the red apple, cook the red potato, cook the yellow apple, cook the yellow potato, fry the green apple, fry the red apple, fry the red potato, fry the yellow apple, fry the yellow potato, grill the green apple, grill the red apple, grill the red potato, grill the yellow apple, grill the yellow potato, roast the green apple, roast the red apple, roast the red potato, roast the yellow apple, roast the yellow potato, cut the cilantro, cut the green apple, cut the parsley, cut the red apple, cut the red potato, cut the yellow apple, cut the yellow potato, chop the cilantro, chop the green apple, chop the parsley, chop the red apple, chop the red potato, chop the yellow apple, chop the yellow potato, dice the cilantro, dice the green apple, dice the parsley, dice the red apple, dice the red potato, dice the yellow apple, dice the yellow potato, slice the cilantro, slice the green apple, slice the parsley, slice the red apple, slice the red potato, slice the yellow apple, slice the yellow potato, eat the cilantro, eat the green apple, eat the parsley, eat the red apple, eat the yellow apple, go to the kitchen, open the cutlery drawer, open the dishwasher, open the fridge, open the kitchen cupboard, open the trash can, You are hungry! Let's cook a delicious meal. Check the cookbook in the kitchen for the recipe. Once done, enjoy your meal!, pick up the cilantro, pick up the cookbook, pick up the green apple, pick up the knife, pick up the parsley, pick up the red apple, pick up the red potato, pick up the yellow apple, pick up the yellow potato.

\section{LLMs Prompts}
\label{app:prompts}
We run all LM calls with OpenAI's \textit{gpt-3.5-turbo} model. We use a temperature of $0$ for the LM Reward Function and a temperature of $0.9$ for the LM Relabeler and the LM Goal Generator. 
\subsection{LM Relabeler Prompt}
\subsubsection{LMA3 $\backslash$ CoT \& Human Tips}
Here is the LM Relabeler prompt with no human tips and no chain-of-thought used for \textit{LMA3 $\backslash$ CoT \& Human Tips}.

\begin{tcolorbox}[fonttitle=\fon{pbk}\bfseries,
                  fontupper=\sffamily,
                  fontlower=\fon{put},
                  enhanced, breakable,
                  title=LMA3 $\backslash$ CoT \& Human Tips]


Exercise: Given the description of a player's behavior in a video game, list the most interesting, impressive, novel or creative goals he achieved and, for each goal, specify when it is achieved for the first time. Write each goal starting with an imperative verb. Here are three examples:\\

Example 1:\\
``````\\
Step 52.\\
Action 52: slice orange bell pepper\\
Observation 52: You slice the orange bell pepper.\\
Step 53.\\
Action 53: cook white onion in oven\\
Observation 53: You roast the white onion with the oven.\\
Step 54.\\
Action 54: cook carrot in stove\\
Observation 54: You fry the carrot with the stove.\\
Step 55.\\
Action 55: cook orange bell pepper in oven\\
Observation 55: You roast the orange bell pepper with the oven.\\
Step 56.\\
Action 56: move north\\
Observation 56: You are in the corridor. In one part of the room you see a shoe cabinet that is closed.There is also a key holder, that has nothing on it. You also see a hat rack, that has nothing on it. In another part of the room you see a coat hanger, that has nothing on it. In one part of the room you see a umbrella stand, that has nothing on it. To the North you see the driveway. To the South you see the kitchen. Through an open wood door, to the East you see the laundry room. Through an open screen door, to the West you see the backyard.\\
"""\\
Let's not forget to mention the steps at which the agent achieved the goal for the first time. Use this format: \{goal description\} (step \{step\_number\}).\\
Answer:\\
- roast a white onion (step 55).\\
- go to the corridor (step 56).\\
- find at the laundry room (step 56).\\
- slice a bell pepper (Step 52).\\
- cook an orange bell pepper (step 55).\\
- find an umbrella stand (step 56).\\

Example 2:\\
``````\\
Step 78.\\
Action 78: open cupboard\\
Observation 78: you open the cupboard, it is empty.\\
Step 79.\\
Action 79: open kitchen drawer\\
Observation 79: you open the kitchen drawer, it contains a knife and a fork.\\
Step 80.\\
Action 80: take knife\\
Observation 80: you pick up the knife\\
Step 81.\\
Action 81: open trash can\\
Observation 81: you open the trash can, it contains an apple\\
Step 82.\\
Action 82: move east\\
Observation 82: You are in the pantry. In one part of the room you see a folding chair that has some olive oil, a roasted red potato, some sugar, some black pepper, and some vegetable oil on it. There is also a shelf that has some salt, and some peanut oil on it. To the West you see the kitchen.\\
Step 83.\\
Action 83: move west\\
Observation 83: You are in the kitchen. In one part of the room you see a stove. There is also an oven. You also see a fridge that is closed. In another part of the room you see a counter, that has a yellow potato on it. In one part of the room you see an open kitchen cupboard, that is empty. There is also a cutlery drawer that is open. To the East you see the pantry.\\
Step 84.\\
Action 84: take yellow potato\\
Observation 84: you take the yellow potato.\\
Step 85.\\
Action 85: slice potato\\
Observation 85: you cut the potato in slices\\
"""\\
Let's not forget to mention the steps at which the agent achieved the goal for the first time. Use this format: \{goal description\} (step \{step\_number\}).\\
Answer:\\
- open the trash can (step 81).\\
- look into the cupboard (step 78).\\
- open the kitchen drawer (step 79).\\
- cut a yellow potato (Step 84).\\
- go the pantry (step 82).\\

Example 3:\\
\textbf{[insert trajectory to relabel here]}\\
Answer:\\
-
\end{tcolorbox}

\subsubsection{LMA3 $\backslash$ Human Tips}
\textit{LMA3 $\backslash$ Human Tips} makes use of chain-of-thought prompting but does not leverage human advice. The example trajectories and answers remain the same as in the previous prompt, but explanations are added before \textbf{each} answer. Here is what both examples from the previous prompt look like.

\begin{tcolorbox}[fonttitle=\fon{pbk}\bfseries,
                  fontupper=\sffamily,
                  fontlower=\fon{put},
                  enhanced, breakable,
                  title=LMA3 $\backslash$ Human Tips - Example 1]

Let's think step by step.\\
Reasoning: Here are some interesting goals the player achieved. The player cooked a white onion (step 53), visited the corridor (step 56), saw the laundry room (step 56), sliced and roasted an orange bell pepper (steps 52 and 55) and saw an umbrella stand (step 56). Let's not forget to mention the steps at which the agent achieved the goal for the first time. Use this format: \{goal description\} (step \{step\_number\}).\\
Answer:\\
- roast a white onion (step 55).\\
- go to the corridor (step 56).\\
- find at the laundry room (step 56).\\
- slice a bell pepper (Step 52).\\
- cook an orange bell pepper (step 55).\\
- find an umbrella stand (step 56).
\end{tcolorbox}

\begin{tcolorbox}[fonttitle=\fon{pbk}\bfseries,
                  fontupper=\sffamily,
                  fontlower=\fon{put},
                  enhanced, breakable,
                  title=LMA3 $\backslash$ Human Tips - Example 2]

Let's think step by step.\\
Reasoning: The agent open various containers: the trash can (step 81), the cupboard (step 78) and the kitchen drawer (step 79). It cut a yellow potato with a knife (step 84) and went to the pantry (step 82). Let's not forget to mention the steps at which the agent achieved the goal for the first time. Use this format: \{goal description\} (step \{step\_number\}).\\
Answer:\\
- open the trash can (step 81).\\
- look into the cupboard (step 78).\\
- open the kitchen drawer (step 79).\\
- cut a yellow potato (Step 84).\\
- go the pantry (step 82).
\end{tcolorbox}

Finally, the end of the prompt include chain-of-thought prompting as well:

\begin{tcolorbox}[fonttitle=\fon{pbk}\bfseries,
                  fontupper=\sffamily,
                  fontlower=\fon{put},
                  enhanced, breakable,
                  title=LMA3 $\backslash$ Human Tips - Example 3]
Example 3:\\
\textbf{[insert trajectory to relabel here]}\\
Let's think step by step and relabel up to 10 goals.\\
Reasoning:
\end{tcolorbox}

\subsubsection{LMA3}
\textit{LMA3} makes use of of both chain-of-thought prompting and human tips. The example trajectories remain the same as in the previous prompts but the answer and reasoning change.

\begin{tcolorbox}[fonttitle=\fon{pbk}\bfseries,
                  fontupper=\sffamily,
                  fontlower=\fon{put},
                  enhanced, breakable,
                  title=LMA3 - Example 1]
                  
Let's think step by step.\\
Reasoning: In the above trajectory, the agent both sliced (step 52) and roasted (step 55) an orange ingredient (orange bell pepper), which demonstrates his capability to prepare an ingredient in several steps. He used the oven twice (steps 53 and 54). He successfully cooked several ingredients: an onion (step 53), the orange bell pepper (step 52) and a carrot (step 54), which shows time-extended commitment to prepare a recipe. An interesting way to describe goals is to mention consecutive steps: here the player first cooked an onion, then cut a bell pepper (the overall goal, made of two steps, is completed in step 55). Interestingly, the player discovered new properties of the environment: he found a place from which he could see both the laundry room and the backyard (from the corridor) in step 56. He found out whether the keyholder hold something in it in step 56 (it did not). Let's not forget to mention the steps at which the agent achieved the goal for the first time. Use this format: \{goal description\} (step \{step\_number\}).\\
Answer:\\
- slice and cook an orange ingredient (step 55).\\
- use the oven for the second time (step 55).\\
- roast an onion and a bell pepper and fry carrots (step 55).\\
- cook an onion first then cut a bell pepper (step 55).\\
- find a place from which you can see both the laundry room and the backyard (step 56).\\
- find out whether the keyholder has something on it (step 56).
\end{tcolorbox}

\begin{tcolorbox}[fonttitle=\fon{pbk}\bfseries,
                  fontupper=\sffamily,
                  fontlower=\fon{put},
                  enhanced, breakable,
                  title=LMA3 - Example 2]

Let's think step by step.\\
Reasoning: In this trajectory, the agent searched for a knife and used it to cut a potato in slices (achieved in step 85). He discovered a new room, the pantry in step 84. He found out that the trash can was not empty (step 81) and looked inside three containers: the trash can (step 81), the cupboard (step 78) and the drawer (step 79). He left the kitchen and came back (step 83). Let's not forget to mention the steps at which the agent achieved the goal for the first time. Use this format: \{goal description\} (step \{step\_number\}).\\
Answer:\\
- find a knife and use it to cut a potato (step 84).\\
- find the pantry (step 84).\\
- open three containers (step 81).\\
- leave and come back to the kitchen (step 83).
\end{tcolorbox}

\subsection{LM Reward Function Prompt}
\subsubsection{LMA3 $\backslash$ CoT}

The prompt of the LM Reward Function without chain-of-thought prompting is the following:

\begin{tcolorbox}[fonttitle=\fon{pbk}\bfseries,
                  fontupper=\sffamily,
                  fontlower=\fon{put},
                  enhanced, breakable,
                  title=LMA3 $\backslash$ CoT]
Exercise: Given the description of a player's behavior in a video game and a list of goals, tell me whether the player achieves these goals and, if he does, when the goal is achieved. Here are three examples.\\

Example 1:\\
``````\\
Step 52.\\
Action 52: slice orange bell pepper\\
Observation 52: You slice the orange bell pepper.\\
Step 53.\\
Action 53: cook white onion in oven\\
Observation 53: You roast the white onion with the oven.\\
Step 54.\\
Action 54: cook carrot in stove\\
Observation 54: You fry the carrot with the stove.\\
Step 55.\\
Action 55: cook orange bell pepper in oven\\
Observation 55: You roast the orange bell pepper with the oven.\\
Step 56.\\
Action 56: move north\\
Observation 56: You are in the corridor. In one part of the room you see a shoe cabinet that is closed.There is also a key holder, that has nothing on it. You also see a hat rack, that has nothing on it. In another part of the room you see a coat hanger, that has nothing on it. In one part of the room you see a umbrella stand, that has nothing on it. To the North you see the driveway. To the South you see the kitchen. Through an open wood door, to the East you see the laundry room. Through an open screen door, to the Wes you see the backyard.\\
"""\\
Here is the list of goals: ``cook an omelet", ``cook an orange ingredient", ``move north, then move south", ``achieved goal: do xx", ``roast two ingredients in the oven", ``cook several ingredients". Let's answer and indicate steps of goal completion:\\
- cook an omelet. Answer: no.\\
- cook an orange ingredient. Answer: yes (step 54).\\
- move north, then move south. Answer: no.\\
- achieved goal: do xx. Answer: no.\\
- roast two ingredients in the oven. Answer: yes (step 55).\\
- cook several ingredients. Answer: yes (step 54).\\

Example 2:\\
``````\\
Step 78.\\
Action 78: open cupboard\\
Observation 78: you open the cupboard, it is empty.\\
Step 79.\\
Action 79: open kitchen drawer\\
Observation 79: you open the kitchen drawer, it contains a knife and a fork.\\
Step 80.\\
Action 80: take knife\\
Observation 80: you pick up the knife\\
Step 81.\\
Action 81: open trash can\\
Observation 81: you open the trash can, it contains an apple\\
Step 82.\\
Action 82: move east\\
Observation 82: You are in the pantry. In one part of the room you see a folding chair that has some olive oil, a roasted red potato, some sugar, some black pepper, and some vegetable oil on it. There is also a shelf that has some salt, and some peanut oil on it. To the West you see the kitchen.\\
Step 83.\\
Action 83: move west\\
Observation 83: You are in the kitchen. In one part of the room you see a stove. There is also an oven. You also see a fridge that is closed. In another part of the room you see a counter, that has a yellow potato on it. In one part of the room you see an open kitchen cupboard, that is empty. There is also a cutlery drawer that is open. To the East you see the pantry.\\
Step 84.\\
Action 84: take yellow potato\\
Observation 84: you take the yellow potato.\\
Step 85.\\
Action 85: slice potato\\
Observation 85: you cut the potato in slices\\
"""\\
Here is the list of goals: ``open an object", ``cook a potato", ``find a knife and cut a potato with it", ``eat a meal". Let's answer and indicate steps of goal completion:\\
- open an object. Answer: yes (step 78).\\
- cook a potato. Answer: no.\\
- find a knife and cut a potato with it. Answer: yes (step 85).\\
- eat a meal. Answer: no.\\

Example 3:\\
\textbf{[insert trajectory here]}\\
Here is the list of goals: \textbf{[insert list of goals to test here]}. Let's answer and indicate steps of goal completion:
\end{tcolorbox}

\subsubsection{LMA3}

With chain-of-thought prompting, we add reasoning description to the answers for each example.

\begin{tcolorbox}[fonttitle=\fon{pbk}\bfseries,
                  fontupper=\sffamily,
                  fontlower=\fon{put},
                  enhanced, breakable,
                  title=LMA3 - Example 1]
- cook an omelet. Reasoning: there is no omelet in this game, this goal is impossible. Answer: no.\\
- cook an orange ingredient. Reasoning: the orange bell pepper and the carrot are two orange ingredients. The carrot is cooked first, in observation 54 so the goal was first achieved in step 54. Answer: yes (step 54).\\
- move north, then move south. Reasoning: the player moves north in step 56, but it does not move south after that. Answer: no.\\
- achieved goal: do xx. Reasoning: this goal does not make sense and thus cannot be achieved. Answer: no.\\
- roast two ingredients in the oven. Reasoning: the player roasts two ingredients in the oven: the white onion (step 53) and the bell pepper (step 55). The goal is only completed in step 55. Answer: yes (step 55).\\
- cook several ingredients. Reasoning: the player cooks a white onion (step 53), a carrot (step 54) and the bell pepper (step 55). The world several requires at least two ingredients, so the goal is completed in step 54 when two ingredients have been cooked. Answer: yes (step 54).
\end{tcolorbox}

\begin{tcolorbox}[fonttitle=\fon{pbk}\bfseries,
                  fontupper=\sffamily,
                  fontlower=\fon{put},
                  enhanced, breakable,
                  title=LMA3 - Example 2]
- open an object. Reasoning: the player opens a cupboard (step 78), a trash can (step 81) and a kitchen drawer (step 79). He achieves the goal for the first time in step (78). Answer: yes (step 78). \\
- cook a potato. Reasoning: the potato is sliced but not cooked. Answer: no.\\
- find a knife and cut a potato with it. Reasoning: the player finds the knife in step 80 and slices a potato in step 84, thus truly completes the goal in step 85. Answer: yes (step 85).\\
- eat a meal. Reasoning: the player does not eat anything here. Answer: no.
\end{tcolorbox}

For the third example, we replace ``\textit{Let's answer and indicate steps of goal completion:}" with ``\textit{Let's think step by step and indicate steps of goal completion:}".

\begin{tcolorbox}[fonttitle=\fon{pbk}\bfseries,
                  fontupper=\sffamily,
                  fontlower=\fon{put},
                  enhanced, breakable,
                  title=LMA3 - Example 3]
Example 3:\\
\textbf{[insert trajectory here]}\\
Here is the list of goals: \textbf{[insert list of goals to test here]}. Let's think step by step and indicate steps of goal completion:
\end{tcolorbox}

\subsection{LM Goal Generator Prompt}
\subsubsection{LMA3 $\backslash$ CoT}
Here is the prompt of the LM Goal Generator without chain-of-thought prompting:

\begin{tcolorbox}[fonttitle=\fon{pbk}\bfseries,
                  fontupper=\sffamily,
                  fontlower=\fon{put},
                  enhanced, breakable,
                  title=LMA3 $\backslash$ CoT]
Context: I am playing a video game, and here is an example of what can happen in that game:\\
\textbf{[insert trajectory here]}\\
Exercise: Using the given list of possible instructions, find a sequence of 2, 3, or 4 instructions that will help me achieve a new, interesting, or creative goal in this game. Do not pick instructions that do not help reaching the main goal, only relevant ones. First describe the new goal starting with an imperative verb; then list the instructions and their corresponding numbers in the list. Here are three examples:\\

Example 1: the list of possible instructions is:\\
\#1 wash the plate\\
\#2 pick up the green apple\\
\#3 pick up the plate\\
\#4 put the potato on the counter\\
\#5 put the plate in the sink\\
Answer: goal: do the dishes. instructions: pick up the plate (\#3); put the plate in the sink (\#5); wash the plate (\#1).\\

Example 2: the list of possible instructions is: \\
\#1 eat the red apple\\
\#2 pick up wood\\
\#3 turn the heat down\\
\#4 pick up an ax\\
\#5 cook an omelet\\
\#6 cut the wood\\
\#7 put the wood in the chimney\\
\#8 turn on TV\\
Answer: goal: prepare a fire in the chimney. instructions: pick up an ax (\#4); pick up wood (\#2); cut the wood (\#6); put the wood in the chimney (\#7).\\

Example 3: the list of possible instructions is: \\
\textbf{[insert subsample of up to 60 mastered subgoals here]}\\
Answer:
\end{tcolorbox}

\subsubsection{LMA3}
With chain-of-thought prompting, we add reasoning to the selection of the main goal and its subgoals.

\begin{tcolorbox}[fonttitle=\fon{pbk}\bfseries,
                  fontupper=\sffamily,
                  fontlower=\fon{put},
                  enhanced, breakable,
                  title=LMA3 - Example 1]
Let's think step by step:\\
Reasoning: You could do the dishes by following less than 4 instructions by first picking up the plate (\#3), then putting it in the sink (\#5), and finally washing the plate (\#1) (3 instructions).\\
Answer: goal: do the dishes. instructions: pick up the plate (\#3); put the plate in the sink (\#5); wash the plate (\#1).\\
\end{tcolorbox}

\begin{tcolorbox}[fonttitle=\fon{pbk}\bfseries,
                  fontupper=\sffamily,
                  fontlower=\fon{put},
                  enhanced, breakable,
                  title=LMA3 - Example 2]
Let's think step by step:\\
Reasoning: You could prepare a fire in the chimney by following 4 instructions. You would need to first pick up an axe (\#4) and pick up wood (\#2), then cut the wood (\#6) and put the wood in the chimney (\#7) (4 instructions).\\
Answer: goal: prepare a fire in the chimney. instructions: pick up an axe (\#4); pick up wood (\#2); cut the wood (\#6); put the wood in the chimney (\#7).
\end{tcolorbox}

For the third example, we replace ``\textit{Answer:}" with a chain-of-thought sentence.

\begin{tcolorbox}[fonttitle=\fon{pbk}\bfseries,
                  fontupper=\sffamily,
                  fontlower=\fon{put},
                  enhanced, breakable,
                  title=LMA3 - Example 3]
Example 3: the list of possible instructions is: \\
\textbf{[insert subsample of up to 60 mastered subgoals here]}\\
Let's think step by step and find an interesting and creative goal to reach:\\
Reasoning:
\end{tcolorbox}

\end{document}